%% file: main.tex
\pdfoutput=1

\documentclass[11pt]{article}

\usepackage[final]{acl}

\usepackage[utf8]{inputenc} 
\usepackage[T1]{fontenc}    
\usepackage{hyperref}       
\usepackage{url}            
\usepackage{booktabs}       
\usepackage{amsfonts}       
\usepackage{nicefrac}       
\usepackage{microtype}      
\usepackage{xcolor}         

\usepackage{times}
\usepackage{latexsym}

\usepackage{utfsym}
\usepackage{fontawesome}
\usepackage{multirow}
\usepackage{adjustbox}
\usepackage{enumitem}
\usepackage{graphicx}
\usepackage{colortbl}
\usepackage{pgf}
\usepackage{pgfplotstable} 
\usepackage{etoolbox}
\usepackage{twemojis}
\usepackage{caption}
\usepackage{pifont}
\usepackage{skak}
\usepackage{subcaption}
\usepackage{soul}

\usepackage[normalem]{ulem}
\useunder{\uline}{\ul}{}

\usepackage{transparent}
\usepackage{amsmath}
\usepackage{amssymb}
\usepackage{cleveref}
\usepackage{accents}

\newcommand{\cmark}{\ding{51}}%
\newcommand{\xmark}{\ding{55}}%
\newcommand{\grayc}[0]{\cellcolor[rgb]{0.957,0.957,0.957}}
\definecolor{mygray}{gray}{0.75}
\newcommand{\g}[0]{\color{mygray}}
\definecolor{myred}{rgb}{0.75, 0.0, 0.0}
\newcommand{\red}[0]{\color{myred}}
\newcommand{\crown}[0]{\includegraphics[height=1em]{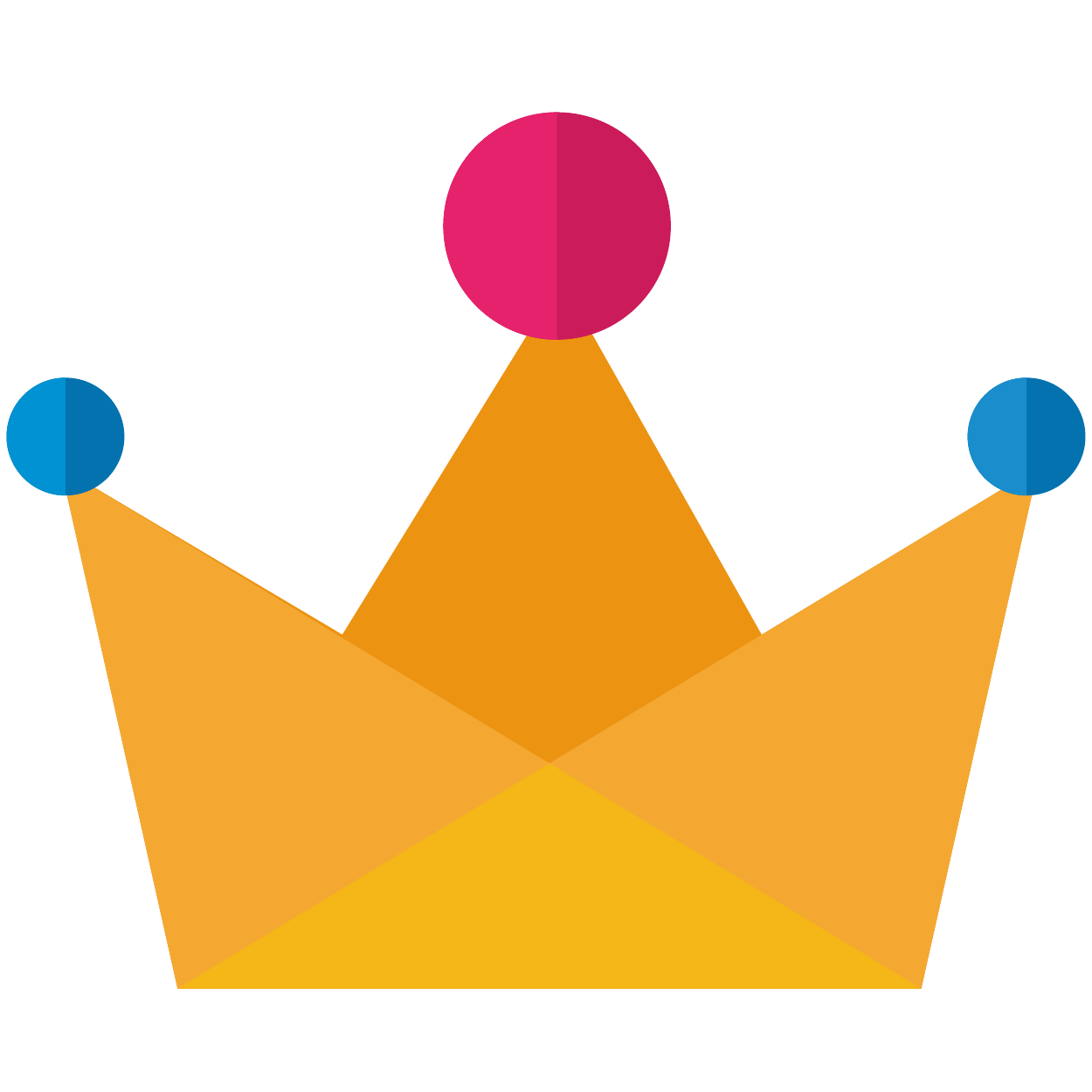}}

\definecolor{eraRule}{HTML}{f0f0f0}
\definecolor{eraSeq2Seq}{HTML}{e0f7fa}
\definecolor{eraTransformer}{HTML}{e8f5e9}
\definecolor{eraLLM}{HTML}{fff3e0}
\newcommand{\graymidrule}{\arrayrulecolor{gray!80}\midrule\arrayrulecolor{black}}

\newcommand{\hlc}[2]{\sethlcolor{#1} \hl{#2}}

\usepackage{listings}
\title{What Really Matters for Table LLMs?\\A Meta-Evaluation of Model and Data Effects}

\author{Naihao Deng$^{1}$\thanks{Work done during Naihao's internship at AWS}, Sheng Zhang$^{2}$, Henghui Zhu$^{3*}$, \textbf{Shuaichen Chang}$^{2}$, \textbf{Jiani Zhang}$^{5*}$,\\
\textbf{Alexander Hanbo Li}$^{4*}$, \textbf{Chung-Wei Hang}$^{2}$, \textbf{Hideo Kobayashi}$^{2}$, \textbf{Yiqun Hu}$^{2}$, \textbf{Patrick Ng}$^{2*}$ \\
$^{1}$University of Michigan \ $^{2}$AWS AI Labs, New York, USA \ $^{3}$Figma \ $^{4}$OKX \ $^{5}$Google\\
$^{1}$\texttt{dnaihao@umich.edu}; $^{2}$\texttt{zshe@amazon.com}; $^{3}$\texttt{hzhu@figma.com} \\
}

\begin{document}

\maketitle

\begin{abstract}
Table understanding has evolved over decades of NLP research.
In this work, we revisit this trajectory and highlight emerging challenges in the LLM era, particularly the paradox of choice: the difficulty of attributing performance gains amid diverse base models and training sets in the context of table instruction tuning.
We replicate four table LLMs by instruction-tuning three foundation models on four existing datasets, yielding 12 models.
We then evaluate these models across 16 table benchmarks. 
Our study is the first to quantitatively disentangle the effects of training data and base model selection, revealing that base model choice plays a more dominant role than the training data itself.
Generalization and reasoning remain challenging, inviting future effort on table modeling.
We open-source our project at \url{https://github.com/dnaihao/table-sft-eacl-2026}.
We open-source our replicated table LLMs at \url{https://huggingface.co/collections/dnaihao/table-llms}.
\end{abstract}

\section{Introduction}
\begin{figure*}[t]
  \footnotesize
  \captionsetup{type=figure}
  \includegraphics[width=\linewidth]{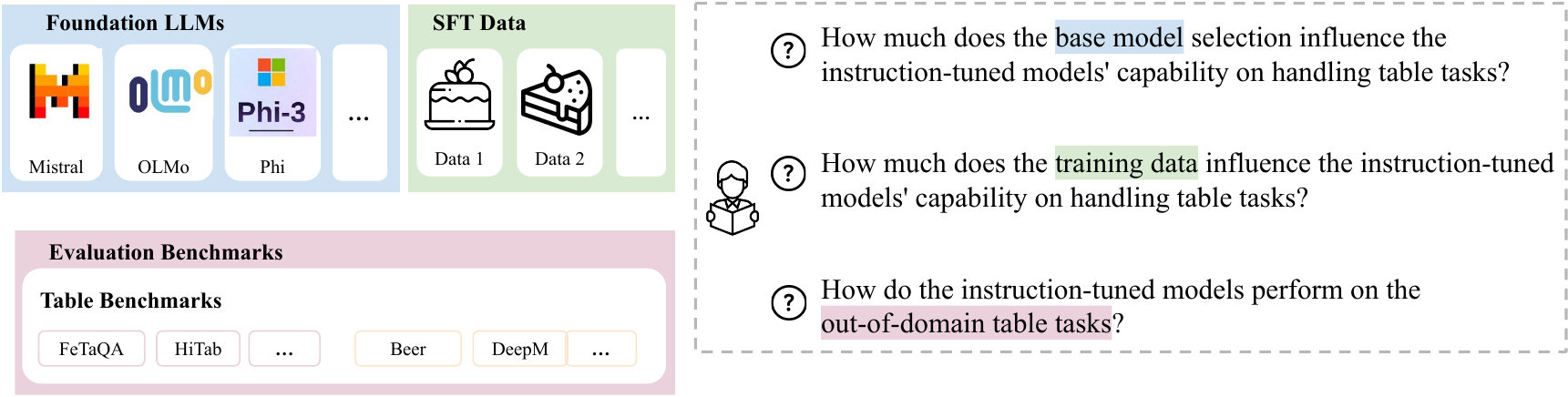}
    \caption{In this paper, we replicate four table LLMs by instruction-tuning three foundation models (OLMo \citep{groeneveld2024olmo}, Mistral \citep{jiang2023mistral}, and Phi \citep{abdin2024phi} models all at 7B scale) on four existing training datasets (TableGPT \citep{li2023table}, TableLlama \citep{zhang-etal-2024-tablellama}, TableLLM \citep{zhang2024tablellm}, TableBench \citep{wu2025tablebench}), yielding 12 models. 
    We evaluate these models across 16 table 
    benchmarks, trying to address the four research questions listed on the right.
    }
\label{fig:concept-diagram}
\end{figure*}

Understanding semi-structured data, such as tables, has been a long-standing challenge in Natural Language Processing (NLP) \citep{woods1972lunar, warren-pereira-1982-efficient, reiter2005choosing, pasupat-liang-2015-compositional, yu-etal-2018-spider, xie-etal-2022-unifiedskg, zhang-etal-2024-tablellama}. 
Over the decades, the field has witnessed a series of paradigm shifts, from symbolic rule-based approaches to neural sequence models, to transformer-based architectures, and now to the era of Large Language Models (LLMs). 
Each shift has come with distinct characteristics and challenges. 
In this paper, we first offer a retrospective framing of these developments and identify the characteristics and challenges associated with table modeling for each era.

The past few years have witnessed a new era for table modeling, characterized by researchers employing instruction tuning for table-specific tasks, giving rise to a wave of ``table LLMs'' \citep{li2023table, zhang-etal-2024-tablellama, zhang2024tablellm, zheng-etal-2024-multimodal, su2024tablegpt2, deng2025rethinking}. 
In the meantime, while the long-standing challenges such as generalization \citep{warren-pereira-1982-efficient, yu-etal-2018-spider, suhr-etal-2020-exploring, deng2025rethinking} and reasoning \citep{liu2018table, xie-etal-2022-unifiedskg, wu2025mmqa} still persist, a new challenge emerges, which we frame as ``paradox of choice''.
Thanks to the numerous foundation LLMs \citep{touvron2023llama, dubey2024llama, jiang2023mistral}, and the diverse table datasets proposed \citep{cheng-etal-2022-hitab, nan-etal-2022-fetaqa}, these table LLMs vary widely in their base model selection, training data, and evaluation datasets. 
With so many moving parts, it has become increasingly difficult to attribute improvements to any one factor, raising concerns about reproducibility and comparability.

In this paper, we select four table LLMs and replicate them by training three distinct foundation LLMs on their proposed dataset, respectively.
While such replication may appear incremental, it directly addresses a blind spot in current table LLM research.
Previous studies \citep{zhang-etal-2024-tablellama, zhang2024tablellm, zha2023tablegpt} typically fix a single base model and attribute progress solely to new instruction data.
However, our controlled experiments reveal that this assumption can be misleading: when the same datasets are applied across three different 7B-scale foundations (Mistral, OLMo, Phi), {\bf base-model choice alone accounts for 81.6\% of performance variance}, whereas {\bf instruction data explains 13.8\%}.
We highlight that this finding turns an intuition long held by practitioners into {\bf quantified, reproducible evidence}, offering the first cross-model, cross-dataset replication for table LLMs.
As a side product, during the replication process, we achieve a new state-of-the-art (SOTA) performance on the HiTab dataset.

Beyond benchmarking, our study introduces a new perspective: {\bf disentangling performance sources} through a unified grid of 12 controlled instruction-tuned models.
This framework enables Shapley–R² attribution analysis, which isolates the effect of foundation quality from that of training data, providing a clearer lens through which to interpret progress in table understanding.
Our study (12 replications × 16 benchmarks) serves as a reminder for practitioners and researchers that scaling up evaluation breadth is just as critical as scaling up model size. 
It highlights the need for broader, standardized comparison practices.

In summary, our contributions are several-fold,
\begin{enumerate}[leftmargin=\parindent,align=left,labelwidth=\parindent,labelsep=0pt]
\item We replicate existing table-LLM setups by instruction-tuning three foundation models on four popular table instruction datasets, yielding 12 directly comparable models.
To the best of our knowledge, this constitutes the first large-scale, cross-model replication in the context of table LLMs.
\item We conduct a comprehensive evaluation across 16 table benchmarks, spanning diverse table-related tasks and generalization scenarios.
This unified evaluation grid provides a consistent basis for comparing instruction-tuning strategies across different foundations.
\item Our analysis highlights the dominant influence of base-model quality on performance — accounting for the majority of observed variance — while showing that current table LLMs continue to struggle with generalization and reasoning.
Together, these findings serve as a reminder for practitioners to scale evaluation scope and methodological rigor.
\end{enumerate}


\section{Backgrounds and Related Works: Paradigm Shift in Table Modeling}
\input{figures/paradigm-shift}

\paragraph{Table-Related Tasks.}
There has been a long history of table-related tasks.
Earlier work has focused on extracting table content from HTML \citep{chen-etal-2000-mining, tengli-etal-2004-learning}.
The deep learning era has seen more diverse table-related tasks such as table question answering (table QA), the task of answering a question given the table and certain context in the format of multiple-choice \citep{jauhar2016tabmcq} and free-form answer \citep{nan-etal-2022-fetaqa}; 
table fact verification, the task of determining whether a given claim is supported or refuted by the table content \citep{Chen2020TabFact, gupta-etal-2020-infotabs}; 
table-to-text, the task of generating a description given the table or some highlighted table cells \citep{parikh-etal-2020-totto};
text-to-SQL, the task of generating a SQL query given the table schema and a user query \citep{zhong2018seqsql, yu-etal-2018-spider}.
These proposed benchmarks cover a diverse set of domains, including Wikipedia tables \citep{parikh-etal-2020-totto}, financial tables \citep{chen-etal-2021-finqa}, scientific tables \citep{moosavi2021scigen}, which serve as invaluable sources for developing and testing general table understanding models.


\paragraph{Paradigm Shift.}
Researchers have explored various methods for table understanding in the past decades, which can date back to the LUNAR system back in the 1970s \citep{woods1972lunar}. 
We briefly summarize the development of table models into four eras (\Cref{fig: table-modeling-history}), where researchers develop rule-based \citep{woods1972lunar, warren-pereira-1982-efficient} and LSTM-based \citep{sutskever2014sequence} algorithms \citep{zhong2018seqsql} in the earlier eras.
With the rise of transformer \citep{vaswani2017attention} and the success of BERT \citep{devlin-etal-2019-bert}, researchers have started to adapt the transformer for table modeling \citep{herzig-etal-2020-tapas,yin-etal-2020-tabert,yu2021grappa,shi2021learning,yang-etal-2022-tableformer}. 
With the success of LLMs \citep{ouyang2022training}, the community has shifted its focus on prompting-based methods \citep{chang2023prompt, deng-etal-2024-tables}\footnote{Since many of the prompting methods are model-agnostic, and we have no information on the model size of the commercial LLMs such as GPT-4, we do not include these methods in \Cref{fig: table-modeling-history}.} as well as instruction tuning the base LLMs \citep{li2023table, zhang-etal-2024-tablellama, zheng-etal-2024-multimodal, zhang2024tablellm}.
\Cref{app-subsec: different-eras} provides additional discussion on the paradigm shifts.


\section{Challenges in Table Modeling}
\label{subsec: challenges-table-modeling}

\begin{table*}[t]
    \small
    \centering
    \renewcommand{\arraystretch}{1.3}
    \setlength\tabcolsep{5pt}
    \resizebox{\linewidth}{!}{
    \begin{tabular}{ccccccccc}
        \toprule
        Model & Base Model & \begin{tabular}[c]{@{}c@{}}Self-Created\\Training Data\end{tabular} & \begin{tabular}[c]{@{}c@{}}Evaluation\\Benchmarks\end{tabular} & \begin{tabular}[c]{@{}c@{}}Open\\Model?\end{tabular} & \begin{tabular}[c]{@{}c@{}}Open\\Data?\end{tabular} & \begin{tabular}[c]{@{}c@{}}Compare w. Other\\ Table LLMs?\end{tabular} & \begin{tabular}[c]{@{}c@{}}Train on Multiple\\ Base LLM?\end{tabular}\\
        \midrule    
        TableGPT (\citeyear{zha2023tablegpt}) & - & - & - & \xmark & \xmark & \xmark & \xmark\\
        \grayc Table-GPT (\citeyear{li2023table}) & \grayc GPT-3.5 & \grayc \cmark & \grayc CTA (\citeyear{deng2022turl}), WikiTQ (\citeyear{pasupat-liang-2015-compositional}), ... & \grayc \xmark & \grayc \cmark & \grayc \xmark & \grayc \cmark \\
        TableLlama (\citeyear{zhang-etal-2024-tablellama}) &  LongLoRA $^\dagger$ & \cmark & FeTaQA (\citeyear{nan-etal-2022-fetaqa}), WikiTQ (\citeyear{pasupat-liang-2015-compositional}), ... & \cmark & \cmark & \xmark & \xmark \\
        \grayc TableLLM (\citeyear{zhang2024tablellm}) & \grayc \begin{tabular}[c]{@{}c@{}}CodeLlama Instruct\end{tabular} & \grayc \cmark & \grayc WikiTQ\textsubscript{m}, TATQA\textsubscript{m}, ... & \grayc \cmark & \grayc \cmark & \grayc \cmark & \grayc \xmark\\
        \begin{tabular}[c]{@{}c@{}}TableBenchLLM (\citeyear{wu2025tablebench})\end{tabular} & \begin{tabular}[c]{@{}c@{}}Llama 3.1 \& others\end{tabular} & \cmark & TableBench (\citeyear{wu2025tablebench}) & \cmark & \cmark & \xmark & \cmark\\
         \bottomrule
    \end{tabular}
    }
    
    \caption{Information for current table instruction tuned models.
    $\dagger$: a variant based on the Llama 2 model.
    We denote the evaluation datasets with a subscript ``m'' as they are adapted by \citet{zhang2024tablellm}.
    We note that these table LLMs are trained from different base LLMs, and each uses its own instruction tuning data, and is tested on a different set of evaluation benchmarks.
    }
    \label{tab: tablellm-comparison}
\end{table*}

There have been challenges for table models in different eras \citep{warren-pereira-1982-efficient, yin-etal-2020-tabert}.
Here, we explain the three challenges we identify for the table LLM era.

\paragraph{Paradox of Choice.}
As we enter the LLM era, a new challenge emerges as the ``paradox of choice'', which refers to the difficulty of choosing from the diverse sets of foundation LLMs and training sets (\Cref{tab: tablellm-comparison}).
We have not seen such a challenge in the previous eras, even in the transformer era, researchers primarily base their models on the BERT model \citep{yin-etal-2020-tabert, herzig-etal-2020-tapas}, and fine-tune their models on a single dataset \citep{yu-etal-2018-spider, wang-etal-2020-rat}.
In contrast, the models in the LLM era adapt different base models \citep{zhang-etal-2024-tablellama, zhang2024tablellm, wu2025tablebench}, some instruction tune these models based on a mix of the existing benchmarks \citep{zhang-etal-2024-tablellama, deng2025rethinking}, while others synthesize their training data \citep{li2023table}.
Such diversified options make it hard to gauge the contributions of base models versus training data in the LLM era, and open up unanswered questions:

\noindent\textit{\textbf{RQ1.}} How much does the base model selection influence the instruction-tuned models' capability on handling table tasks?

\noindent\textit{\textbf{RQ2.}} How much does the training data influence the instruction-tuned models' capability on handling table tasks?

\paragraph{Generalization.}
While table LLMs demonstrate competitive performance \citep{zhang-etal-2024-tablellama}, whether they pick up the table understanding capabilities or overfit to the dataset-specific patterns is still debatable \citep{deng2025rethinking} and opens up a research question:

\noindent\textit{\textbf{RQ3.}} How do the instruction-tuned models perform on the out-of-domain table tasks? 





\Cref{app-sec: challenges-in-table-modeling} provides additional discussion.

\section{Experimental Setups}
\label{sec: experimental-setups}

Because of the limited computing resources and non-trivial computational costs to train and test LLMs, we cannot exhaust all possible evaluations.
\paragraph{Model Selection.}
To rigorously study the influences of base model selection and training data, we select three LLMs that are all released in the year of 2023 and 2024 from non-profit organizations or companies, Mistral-7B-Instruct-v0.3 \citep{jiang2023mistral}, OLMo 7B Instruct \citep{groeneveld2024olmo} and Phi 3 Small Instruct (7B) \citep{abdin2024phi} as our base models, detailed in \Cref{app-sec: experimental-setup}.
\paragraph{Replication.}
For each base model, we replicate the instruction tuning stage for TableLlama \citep{zhang-etal-2024-tablellama}, TableLLM \citep{zhang2024tablellm}, TableBenchLLM \citep{wu2025tablebench}, and Table-GPT \citep{li2023table}.
Our implementation yields comparable or better results than the performance reported in the existing works (\Cref{fig:in-domain-perf}, additional details in \Cref{app-sec: reimpl-compare}). 
\paragraph{Evaluation.}
We select eight real-world table understanding datasets, eight synthetic table understanding datasets
(details in \Cref{app-sec: eval-benchmarks}) for our evaluation.
We note that our controlled replication enables an apples-to-apples comparison and allows us to disentangle the respective contributions of base model capabilities and instruction tuning datasets, therefore better answering the research questions we propose in \Cref{subsec: challenges-table-modeling} (\Cref{{fig:concept-diagram}}).


\section{Results and Discussions}

\input{figures/in-domain-avg}
\input{tables/out-of-domain-complete}

\Cref{fig:in-domain-perf} presents the averaged in-domain (ID) performance.
\Cref{tab: out-of-domain-complete} presents the out-of-domain (OOD) evaluation on various table understanding benchmarks.

\paragraph{RQ1: How much does the base model selection influence the instruction-tuned models' capability on handling table tasks?}

\begin{figure}[t]
    \centering
    \begin{subfigure}[t]{0.35\textwidth}
    \includegraphics[width=0.95\linewidth]{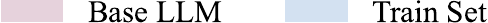}
    \end{subfigure}
    \vspace{0.3em}
    
    \begin{subfigure}[t]{0.45\textwidth}
    \includegraphics[width=0.95\linewidth]{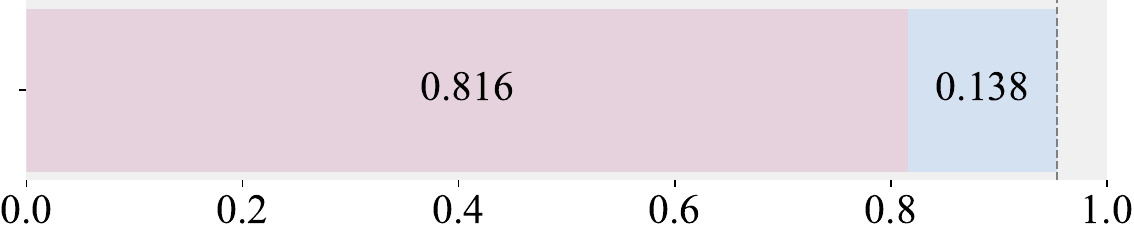}
    \end{subfigure}
    \caption{Shapley R$^2$ decomposition \citep{shapley1953value, israeli2007shapley} for the contributions of the downstream tasks' performance by the base LLM versus the training set.
    We can see that the choice of the base LLM is a dominant factor (0.816 compared to 0.138 from the train set) that decides the model's performance on downstream tasks.
    \Cref{fig:shapley} provides the additional analysis for pair-wise base model comparisons.
    }
    \label{fig:shapley-part}
\end{figure}

\paragraph{Answer: Large OOD performance variance across base models.}
Contrary to performance in \Cref{fig:in-domain-perf}, where we see minimal ID performance variance across different base models, there is a large performance variance across different base models on the OOD table tasks, as shown in \Cref{tab: out-of-domain-complete}.
For instance, when all trained on TableBenchLLM, Phi achieves 83.0 on TabMWP, significantly outperforming Mistral (70.6) and OLMo (62.6). 

\paragraph{The base model is crucial, and in some cases, a determinant factor for the OOD performance.}

In \Cref{fig:shapley-part}, we employ the Shapley R$^2$ decomposition to decompose the performance contributions of the base LLM selection versus the different instruction tuning data (additional details in \Cref{app-subsec: shapley-decomp}).
We find that the base LLMs' selection holds an R$^2$ of 0.816, significantly larger than 0.138, the share of the instruction tuning data.
The share for the base LLM selection remains crucial when we consider model pairs in \Cref{fig:shapley} in \Cref{app-subsec: shapley-decomp}, suggesting that the base model selection is a non-negligible, and sometimes a dominant factor that determines the instruction-tuned model's performance.
However, existing works for table instruction tuning \citep{li2023table, zhang-etal-2024-tablellama, zhang2024tablellm, su2024tablegpt2} barely provide such comparison studies, and typically train their models from a single base LLM, ignoring the crucial factor of base model selection. 

\begin{figure}[t]
        \centering
    \includegraphics[width=0.95\linewidth]{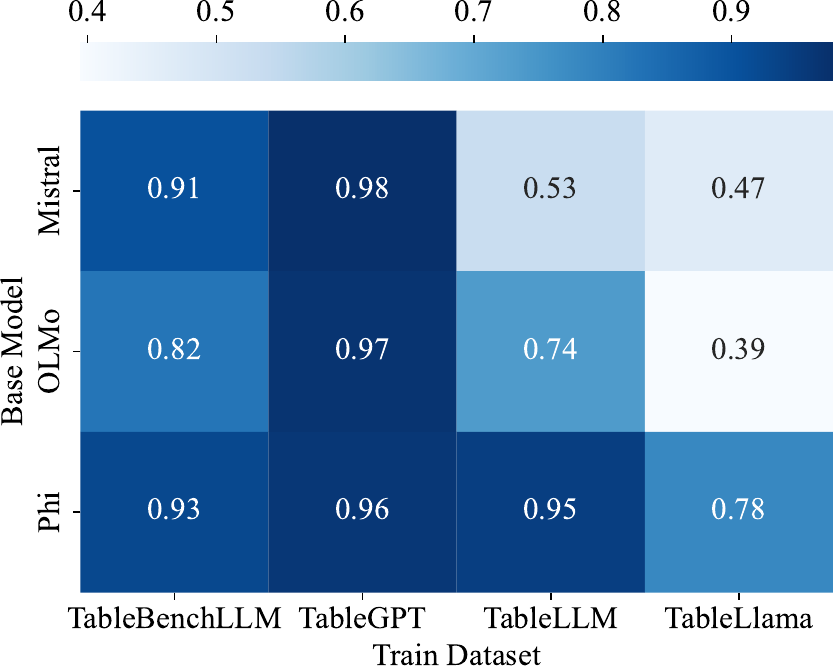}
       \caption{Pearson r scores for the fine-tuned model's performance v.s. the base model's performance on the OOD datasets.
    We find that in general, there is a strong linear correlation between the two performances, with a Pearson r of around 0.7 to 0.9. 
    Even the lowest Pearson r score, 0.39, indicates a moderate positive correlation.}
    \label{fig:pearson-correlation}
\end{figure}

\paragraph{Strong base model leads to significantly better OOD performance.}
In \Cref{fig:pearson-correlation}, we plot the Pearson r scores for the instruction-tuned model's performance v.s. the base model's performance on the out-of-domain datasets.
In general, there is a strong linear correlation between the two performances (Pearson r around 0.7 to 0.9), suggesting that the instruction-tuned model's performance is strongly related to the base model's performance on these table tasks.
We notice that in \Cref{tab: out-of-domain-complete}, the best performance for a single dataset is typically achieved by fine-tuning the Phi model.
We note that the Phi model consistently outperforms the other two models even when untuned.
For instance, TabMWP's overall best performance is achieved by fine-tuning the Phi model with the TableBench training data, and the original Phi model achieves 76.1, outperforming the original Mistral's 66.9 and the original OLMo's 54.4.
TATQA's overall best performance is achieved by fine-tuning the Mistral model with TableBench training data, and the original Mistral model achieves 18.0, outperforming the original OLMo's 14.3 and the original Phi's 13.0.
This suggests that while instruction tuning can meaningfully improve a model’s performance on table tasks, its effectiveness is still heavily bound by the capabilities of the underlying base model.


\begin{figure}[t]
    \centering
    \includegraphics[width=0.9\linewidth]{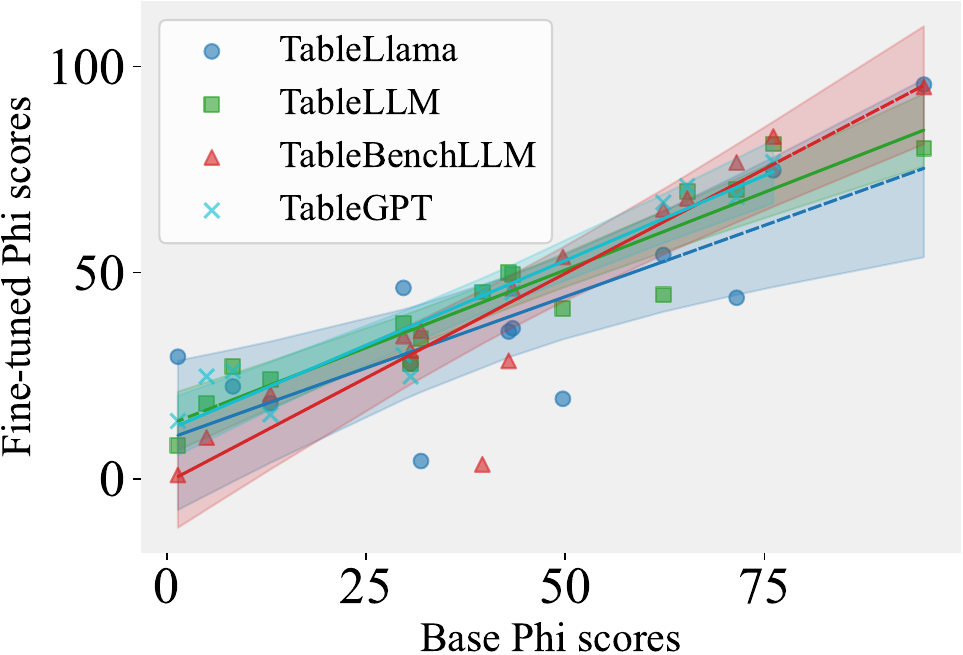}
    \caption{Fine-tuned models' performance (y-axis) with respect to each training dataset v.s. the base Phi model's performance (x-axis) on the OOD table datasets.
    We find that there is a linear correlation (Pearson r ranges from 0.78 to 0.96) between these two scores.}
    \label{fig:phi-data-correlation}
\end{figure}


\paragraph{RQ2. How much does the training data influence the instruction-tuned models' capability in handling table tasks?}

\paragraph{Answer: Instruction tuning yields a significant performance boost for ID datasets.} 
When the dataset is included as part of the training set (e.g., FeTaQA in TableLlama), we observe a significant performance boost compared to the untrained base model (Mistral trained on TableLlama data achieves 38.7 compared to the base's 20.0 on FeTaQA).
This echoes with the finding by \citet{zhang-etal-2024-tablellama, deng2025rethinking} that instruction tuning can significantly boost the ID performance.

\definecolor{col1}{HTML}{fd7f6f}
\definecolor{col2}{HTML}{7eb0d5}
\definecolor{col3}{HTML}{b2e061}
\definecolor{col4}{HTML}{bd7ebe}
\definecolor{col5}{HTML}{ffb55a}
\definecolor{colt1}{HTML}{fed3cd}
\definecolor{colt2}{HTML}{c5dbec}
\definecolor{colt3}{HTML}{cbea95}
\definecolor{colt4}{HTML}{e5cce5}
\definecolor{colt5}{HTML}{ffddb3}
\begin{table*}[t]
    \centering
    \small
    \begin{tabular}{cp{1in}p{1.4in}p{3.1in}}
    \toprule
    \multicolumn{2}{c}{Error Types}  & \multicolumn{1}{c}{Description} & \multicolumn{1}{c}{Example} \\\midrule

    {\color{col3}$\blacktriangleright$} & \emph{Grounding Error}
    & Fail to properly attend to the correct information.
    & 
    \includegraphics[width=10pt]{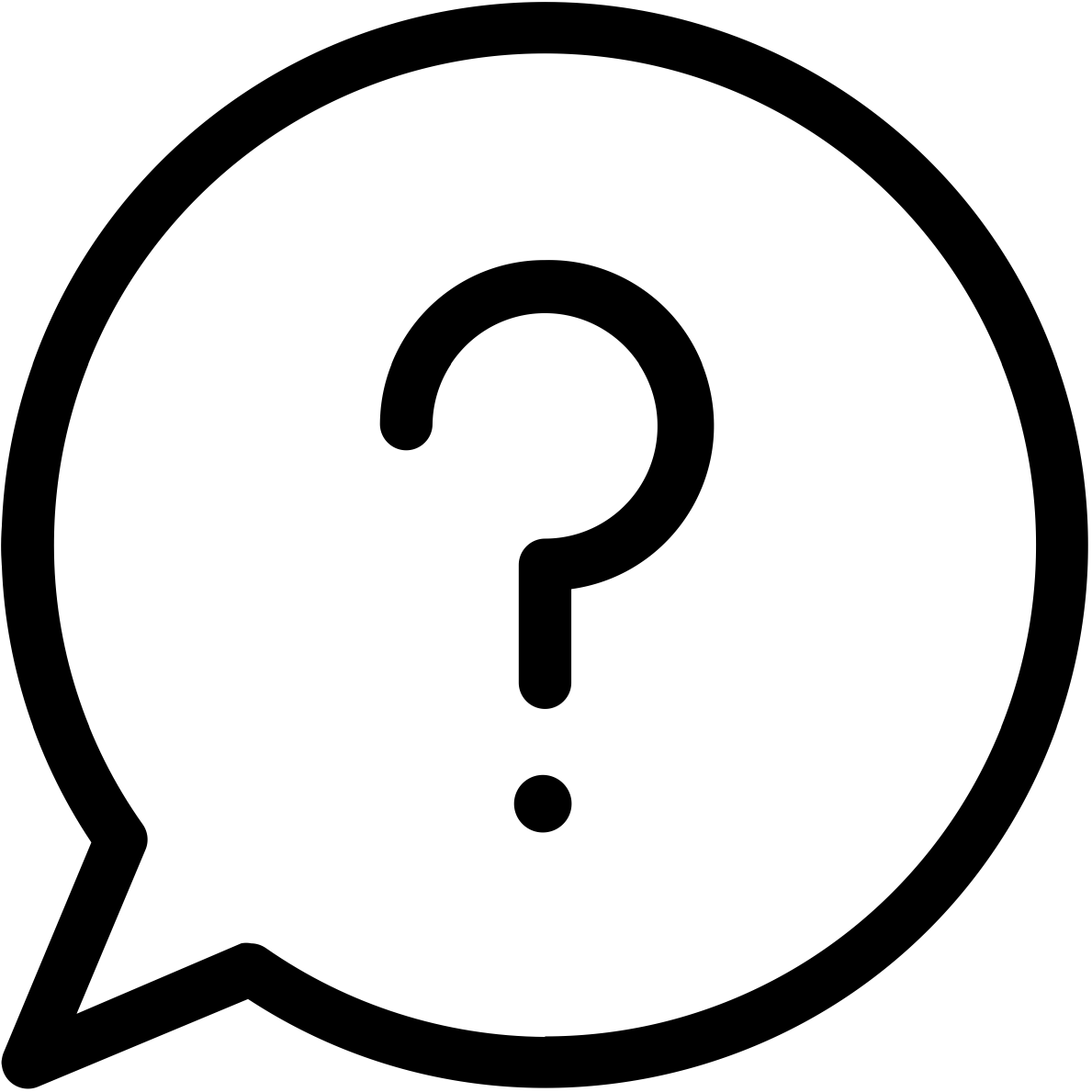} : Find the column that contains the cell value ``348.55''.  
    
    \vspace{0.5em}
    \begin{tabular}{|c|c|c|c|}
    \hline
      $\dots$ &  BalanceLeftTD   & Current Month & $\dots$ \\
      \hline
      $\dots$ & 48796.94   & 348.55 & $\dots$\\
      \hline
    \end{tabular}
    \newline
    \includegraphics[width=10pt]{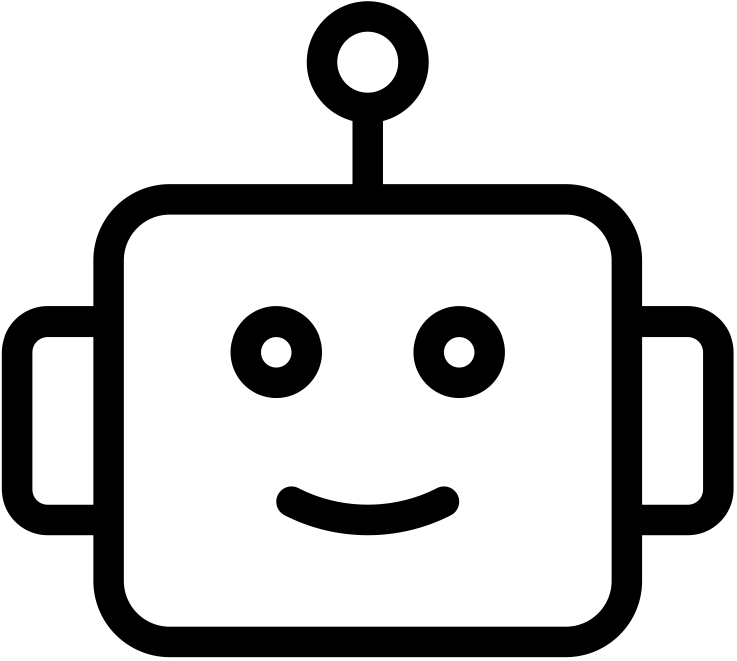} : \hlc{colt3}{BalanceLeftTD} 
    \\\midrule

    {\color{col4}$\blacktriangleright$} & \emph{Math Reasoning Error}
    & Fail to conduct the math reasoning process correctly.
    & \includegraphics[width=10pt]{imgs/icons/robot3.pdf} : $\dots$ the Soviet Union received 29 medals, while East Germany received 25 medals. Therefore, the Soviet Union \hlc{colt4}{did not receive 4 more medals} than East Germany$\dots$ 
    \\\midrule

    {\color{col1}$\blacktriangleright$} & \emph{Not Following Instructions}
    &  Generate output while not following the instruction.
    & \includegraphics[width=10pt]{imgs/icons/question.pdf} : $\dots$ Let's think step by step and show your reasoning before showing the final result $\dots$  \newline
    \includegraphics[width=10pt]{imgs/icons/robot3.pdf} : \hlc{colt1}{Answer: No}\\\midrule
    {\color{col5}$\blacktriangleright$} & \emph{Hallucination}
    & Fabricate ungrounded details or facts.
    & (In the table, Canada has 3 bronze medals; Switzerland has 5.)
    \newline
    \includegraphics[width=10pt]{imgs/icons/robot3.pdf} : $\dots$ According to the table, Switzerland (SUI) and Canada (CAN) \hlc{colt5}{both received 3 bronze medals} $\dots$\\\midrule
    {\color{col2}$\blacktriangleright$} & \emph{Commonsense\newline Errors}
    & Generate outputs that violate common sense.
    & \includegraphics[width=10pt]{imgs/icons/robot3.pdf} : $\dots$ 
    release date is November 11, 2008. However, it \hlc{colt2}{does not provide any information} about the season in which it was released. 
    Therefore, $\dots$
    \\

    \bottomrule
    \end{tabular}
    \caption{Types of reasoning errors commonly made by tableLLMs, with their description and example erroneous responses (\includegraphics[width=10pt]{imgs/icons/robot3.pdf}) to questions (\includegraphics[width=9pt]{imgs/icons/question.pdf}) from our experiment results on the Phi model trained on TableLLM data.}
    \label{tab:error-example}
\end{table*}

\paragraph{Certain training data consistently yield the best OOD performance across different base LLMs.}

Though in \Cref{fig:shapley}, compared to the base LLM selection, the influence of the existing training data remains small in most cases, there is still a linear relation between the training set selection and the instruction-tuned model's performance, as illustrated in \Cref{fig:phi-data-correlation}
In addition, we notice that TableLLM's training data consistently achieves the best (e.g., on HiTab or competitive performance on table QA tasks across all three base models in \Cref{tab: out-of-domain-complete}.
In contrast, though the recipe for TableLlama's training data contains table QA tasks, models trained with the training data from TableLlama underperform those from TableLLM.
We attribute the effectiveness of TableLLM's training data on the table QA task to that when constructing the data, \citet{zhang2024tablellm} leverage LLMs such as GPT-3.5 to enhance the reasoning process (more in \Cref{appx-subsec: train-data-examples}).
Such an enhanced reasoning path would benefit the model's reasoning process, as suggested by the findings by \citet{guo2025deepseek, muennighoff2025s1}.

\paragraph{RQ3. How do the instruction-tuned models perform on the OOD table tasks? }

\paragraph{Answer: The best OOD performance is significantly below the ID performance.}
As shown in \Cref{tab: out-of-domain-complete}, though there are improvements from the base models on the OOD table tasks, the models' performance is far below that of the ID tuned models.
For instance, for the Phi model, if the training set includes HiTab, the model achieves 63.6 (the gray value in \Cref{tab: out-of-domain-complete}), while the best OOD performance on HiTab is 45.3 (achieved by training the Phi model using TableLLM's training set).
Such a large performance gap suggests a large space for improvement.

\paragraph{The instruction-tuned model may yield worse performance than the base model.}
We note that instruction tuning sometimes leads to decreased OOD performance compared to the base model.
For instance, the untuned Mistral model achieves a score of 27.9 on WikiTQ, whereas instruction-tuning it on TableGPT data reduces performance to 25.5. 
This highlights a potential trade-off introduced by instruction tuning.
While it improves alignment on in-domain tasks, it may also cause the model to overfit or overspecialize, leading to reduced generalization on unseen tasks.


\begin{figure}[t]
    \centering
    \includegraphics[width=\linewidth]{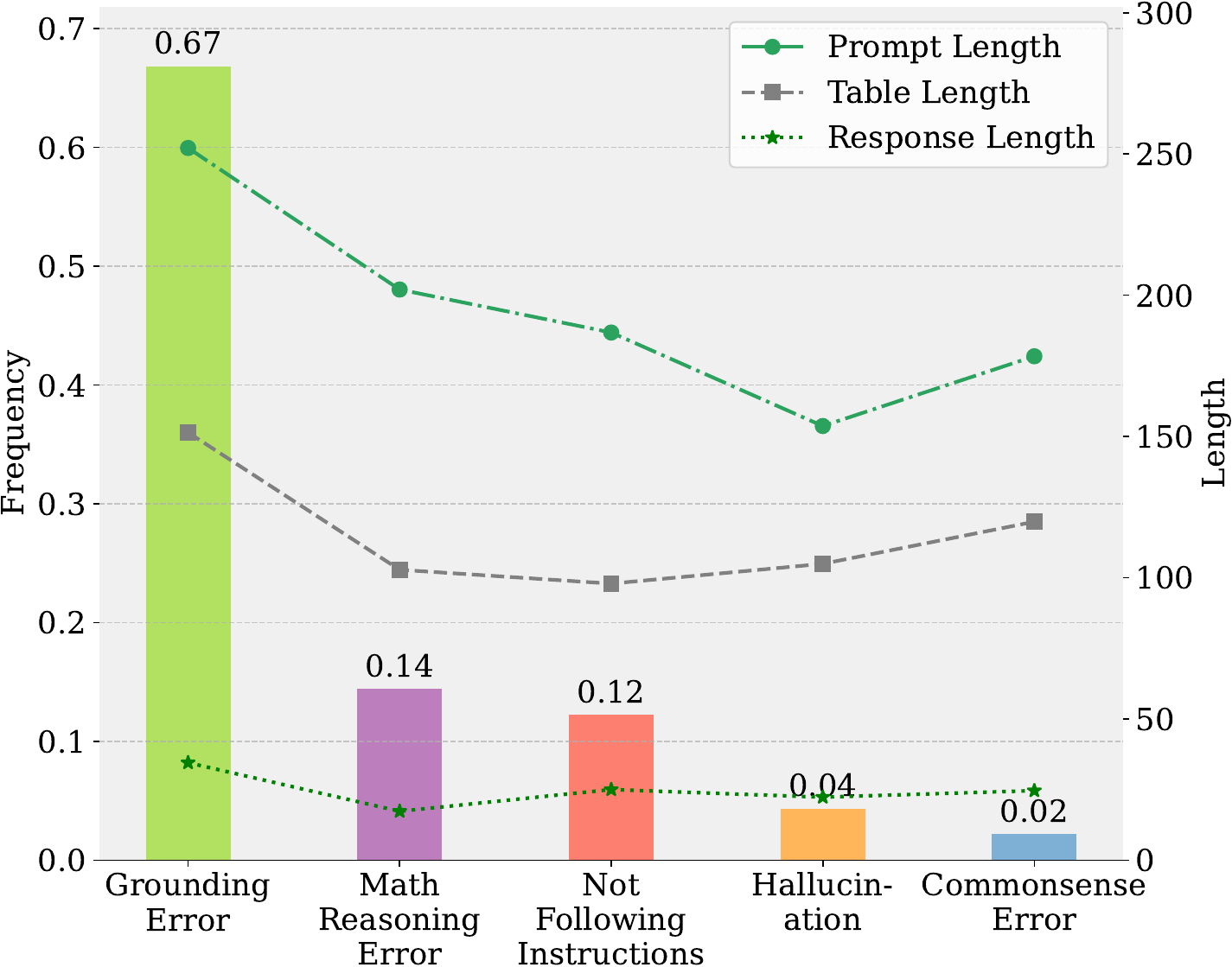}
    \caption{Frequencies of the TableLLM's answers containing the five
reasoning errors, and the corresponding prompt, table, and response length.}
    \label{fig:error-frequency}
\end{figure}

\paragraph{Instruction-tuned models are still prone to grounding and numerical reasoning errors.}
We conduct an error analysis on 1,000 samples predicted by the Phi model fine-tuned on TableLLM data to understand its error cases. 
Representative error cases and their distribution are shown in \Cref{tab:error-example} and \Cref{fig:error-frequency}, respectively.
We find that grounding errors of failing to correctly associate the question with the relevant table content, are the most frequent, particularly in examples involving longer tables or prompts. 
This suggests that instruction tuning alone may be insufficient to develop robust table grounding capabilities, highlighting the need for future work focused on improving models' alignment with tabular inputs.
In addition, models frequently struggle with basic numerical reasoning, such as subtraction over table entries. 
This suggests a persistent limitation in integrating arithmetic operations in the context of table understanding.
Moreover, we observe instruction-following failures in certain cases, aligning with prior findings that further instruction tuning may degrade the base model’s inherent capabilities \citep{wang2023far}. 
While hallucinations and commonsense errors also occur, they are relatively less frequent in table-based tasks compared to general benchmarks \citep{clark2018think, rein2023gpqa}.

\paragraph{Additional RQs and Discussions.}
In addition, we explore several auxiliary research questions and describe our findings here.
First, we find that table instruction tuning does not necessarily compromise general capabilities (\Cref{app-subsec: affect-general-capabilities}).
Second, scaling model size yields diminishing returns under fixed data and compute budgets — larger models improve performance but at significant computational cost (\Cref{app-subsec: model-size}).
Third, when compared to proprietary LLMs, the best-performing open models (e.g., Phi-based variants) narrow the gap on structured reasoning benchmarks, underscoring the progress of open LLMs (\Cref{app-subsec: compare-propriety-llms}).
Finally, we include results on additional datasets that confirm the same trends: base model quality remains the dominant explanatory factor across both real and synthetic table tasks (\Cref{app-subec: addi-datasets}).

\section{Take-Aways}

\paragraph{Base Model Selection Matters Most.}
Our analysis confirms that base model choice is the single most influential factor driving downstream performance. 
As shown in \Cref{fig:shapley-part,fig:pearson-correlation}, instruction-tuned performance correlates strongly with the base model’s intrinsic capability, with base-model factors explaining over 80\% of the observed variance.

\paragraph{Instruction Tuning Helps—but Its Impact Is Often Overstated.}
While instruction tuning substantially improves in-domain performance, its impact on out-of-domain generalization remains limited. 
Thus, much of the leaderboard gain observed in prior work likely reflects the strength of the underlying foundation model rather than improvements from the new instruction data.

\paragraph{Persistent Weaknesses in Generalization and Reasoning.}
Despite overall gains, table LLMs continue to exhibit fragile reasoning behaviors, especially in grounding (linking textual queries to table content) and numerical reasoning (handling arithmetic or comparison). 
Models tend to show sharp performance drops under domain shift.

\paragraph{Toward More Reproducible Evaluation.}
Taken together, these findings highlight the need for systematic, cross-model evaluation frameworks that can disentangle data, model, and training-driven effects.
Scaling evaluation breadth, not just model size, will be essential for ensuring that future progress in table modeling is interpretable, reproducible, and scientifically grounded.
We provide discussions on the future directions for table modeling in \Cref{app-subsec: future-directions}.

\section{Conclusion}
In this paper, we revisit the trajectory of table instruction tuning and re-examine what truly drives progress in table understanding. 
By systematically replicating four popular table LLM setups across three distinct foundation models, we disentangle the intertwined effects of base-model quality and instruction data. 
Our large-scale experiments, covering twelve instruction-tuned models and sixteen evaluation benchmarks, provide the first controlled analysis that quantifies their relative contributions. 
The results reveal a striking imbalance: while instruction data plays a meaningful role, base model selection alone explains over eighty percent of performance variance. 
This finding turns an intuitive belief among practitioners into reproducible evidence and calls for a rethinking of how progress in table modeling is measured.



In revisiting the foundations of table instruction tuning, we highlight that what appears as empirical progress often reflects underlying disparities in model strength.
Recognizing and quantifying these effects is not merely diagnostic—it is essential for building a reliable science of table understanding. 
Future efforts should extend this controlled evaluation framework to broader model scales and modalities, establishing reproducibility as a central norm in table-LLM research.


\section*{Limitations}

We believe our work presents the first of its kind large-scale controlled analysis that explicitly decouples the effects of base model and instruction tuning data in the table understanding domain.
In addition, we want to stress the massive training effort we have invested in, as noted in \Cref{sec: experimental-setups}.
As a side product, we have achieved the new SOTA performance on the HiTab dataset, and provide the first open-source model replication of existing closed-source table LLMs such as Table-GPT.
Moreover, we have comprehensively evaluated these twelve models on 16 table understanding benchmarks.

However, there exist other base models, or other datasets proposed by the researchers, which can be used to train the table LLMs and evaluate these models' capabilities, and by no means can we exhaust all of them in this paper.
We encourage future efforts in comprehensively evaluating these table LLMs' capabilities, and we believe our work has laid a solid foundation for decoupling the contributions of training data and base models, and further enhancing our understanding of table instruction tuning.

While our study provides the first systematic comparison across multiple table LLMs, we acknowledge that only three base models were examined. 
Although our results provide strong empirical evidence, further validation across larger models remains valuable.”
We expect our study to remind practitioners of 

Given the computational cost, we limited our replication to three representative 7 B-scale models. 
Although this covers a diverse spectrum (open, instruction-tuned, and academic), we have no way to obtain results from proprietary LLMs (e.g., commercially closed-source LLMs).
We provide additional discussions on comparing these instruction-tuned table LLMs to proprietary LLMs in \Cref{app-subsec: compare-propriety-llms}.

\section*{Ethical Considerations}
In this work, we isolate the contributions of training data proposed by the existing table LLMs by training the same base models and comparing their performance.
The base models we have used in this work include Mistral v0.3 7B Instruct model \citep{jiang2023mistral}, OLMo 7B Instruct \citep{groeneveld2024olmo}, and Phi 3 Small Instruct (7B) \citep{abdin2024phi}.
We conduct additional studies on Phi 3 Mini Instruct (4B) in \Cref{app-sec: results-and-discussions}.
Foundational models like Mistral v0.3 7B Instruct model are susceptible to jail-breaking instructions \citep{wei2024jailbroken} and may lead to harmful behaviors.
Our objective in this work is to understand the limitations of the existing table instruction tuning, and we urge practitioners to stick to the good purpose when developing or using our models.
Our replicated models can serve as baseline models for future research on structured data, and we provide a holistic evaluation of these models on both table tasks and how they compromise their general capabilities.
Our results lead to various findings on what training data helps the models most on these table tasks, and how to construct LLMs specialized in tables efficiently.

\bibliography{aclanthology,custom}

\newpage
\appendix
\newpage

\section{Backgrounds and Related Works: Paradigm Shift in Table Modeling}
\label{app-sec: table-modeling-history}

\subsection{Captions of \Cref{fig: table-modeling-history}}
\label{app-subsec: table-model-history-fig-explain}
In \Cref{fig: table-modeling-history}, we use ``+'' and ``++'' to denote different sizes of the same model. 
For instance, TAPAS \citep{herzig-etal-2020-tapas} refers to the model based on the small version of the BERT-base model, while TAPAS+ refers to the model based on the large version of the BERT-base model.
For the LSTM models such as \citet{liu2018table}'s model, we estimate the parameter sizes based on the description in the original paper.

\subsection{Different Eras for Table Modeling}
\label{app-subsec: different-eras}
Here we provide further discussions on different eras for table modeling.
\paragraph{Rule-Based and Seq2Seq Era.} The first era is characterized by the symbolic and static nature of the proposed algorithms \citep{woods1972lunar, warren-pereira-1982-efficient}.
Later, with the rise of LSTM in NLP \citep{sutskever2014sequence}, researchers have incorporated domain-specific features into the models, such as specific components to generate SQL queries to query database tables \citep{zhong2018seqsql}.
\paragraph{Transformer Era.}
The earlier trend of domain-specific feature engineering from seq2seq era has made its way into the transformer era, where the pre-trained transformer models \citep{vaswani2017attention}, such as BERT \citep{devlin-etal-2019-bert} have taken over most fields in NLP.  
\citet{herzig-etal-2020-tapas} incorporate embeddings designed for rows and columns, 
\citet{yang-etal-2022-tableformer} adapt the attention mechanism to better align with table structures.
In addition, this era has witnessed a trend of domain-specific pre-training, where researchers collect a large table pre-training corpus \citep{yin-etal-2020-tabert} and design table-specific training objectives \citep{yu2021grappa, shi2021learning}.
\paragraph{LLM Era.}
Ever since the successful launch of the ChatGPT system \citep{ouyang2022training}, researchers have increasingly focused on adapting LLMs for table tasks.
As LLMs have inherent abilities on table understanding, researchers employ prompt engineering on these LLMs for better performance on tables \citep{chang2023prompt, deng-etal-2024-tables}\footnote{Since many of the prompting methods are model-agnostic, and we have no information on the model size of the commercial LLMs such as GPT-4, we do not include these methods in \Cref{fig: table-modeling-history}.}.
Another line of research involves instruction-tuning LLMs by adapting existing table-related benchmarks.
This leads to various table LLMs such as Table-GPT \citep{li2023table}, TableLlama \citep{zhang-etal-2024-tablellama}, TableLlava \citep{zheng-etal-2024-multimodal}, and TableLLM \citep{zhang2024tablellm}. 
Recently, with the rise of reasoning models \citep{guo2025deepseek}, researchers have leveraged reinforcement learning algorithms such as GRPO \citep{shao2024deepseekmath} for table modeling \citep{yang2025table, wu2025table}.
In this paper, we specifically focus on table instruction tuning, a popular post-training paradigm that aligns large language models with structured reasoning tasks over tabular data.

\paragraph{\textit{Remarks.}}
We have seen continuous efforts that last several decades where researchers adapt general modeling methods to the domain of table understanding.
As a result, much like the trend in the general language models, there has been a logarithmic increase in terms of the table model size in the past decades (\Cref{fig: table-modeling-history}).
While these models have kept pushing the state-of-the-art performance on many benchmarks \citep{zhang-etal-2024-tablellama}, the monotonic increase in model sizes is concerning as it limits the access for many research labs where there is no abundant computing resources.

\section{Challenges in Table Modeling}
\label{app-sec: challenges-in-table-modeling}

In the rule-based era, crafting the rules can be labor-intensive \citep{warren-pereira-1982-efficient}; in the transformer era, crafting a large-scale pre-training corpus is data-intensive \citep{yin-etal-2020-tabert}.
In addition to the discussion in \Cref{subsec: challenges-table-modeling}, here we further discuss the generalization. 

\paragraph{Generalization.}
The challenge of generalization has shifted across eras.
Since the rules in earlier systems are hand-crafted and static, the challenge lies primarily in handling the cases where their rules do not cover \citep{warren-pereira-1982-efficient}.
Such problems are mediated with the appearance of the learning-based models (e.g., LSTM, transformers), where the models may have a chance to conduct compositional reasoning to generalize to unseen examples \citep{zhong2018seqsql}.
However, an LSTM model excelled on one domain may fail on other domains \citep{yu-etal-2018-spider}.
This persists in the transformer era, where models perform well on one dataset demonstrate near-zero performance on others \citep{suhr-etal-2020-exploring}.
While in the table LLM era, there seem to be some promises on generalization to unseen tasks \citep{zhang-etal-2024-tablellama}, in our paper, we reveal that generalization challenges remain.

\section{Experimental Setup}
\label{app-sec: experimental-setup}

\paragraph{Foundation LLM Selections.}
For the training data from each existing work, we fine-tune Mistral-7B-Instruct-v0.3 \citep{jiang2023mistral}, OLMo 7B Instruct \citep{groeneveld2024olmo} and Phi 3 Small Instruct (7B) \citep{abdin2024phi}.
Following \citet{zhang-etal-2024-tablellama, zhang2024tablellm, wu2025tablebench}, we fine-tune all the models through full parameter fine-tuning.

\paragraph{Hyperparameter Selection.}
To rule out the effects of the learning rate, we train all three models using a set of learning rates: 5e-5, 1e-5, 5e-6, 1e-6, 5e-7, 1e-7, 5e-8, and 1e-8.
Empirically, we find that they achieve the best when the learning rate is 5e-7.
We do not see significant performance changes as we increase the training steps.
For consistency, we fine-tune our models for three epochs across all the experiments.

We run our experiments on 1 server node with 8 A100, each with 48 GB GPU memory.
We set the batch size to 16 in our training process.

\section{Evaluation Setups}
\label{app-sec: eval-benchmarks}
\subsection{Real-World Table Understanding Benchmarks.}
\paragraph{Dataset Description.}
We evaluate our replicated models on eight existing real-world datasets covering the tasks of table question answering (table QA), table fact verification, and table-to-text generation.
\textbf{FeTaQA (FeT)} \citep{nan-etal-2022-fetaqa} is a free-form table QA dataset sourced from Wikipedia-based tables.
\textbf{HiTab (HiT)} \citep{cheng-etal-2022-hitab} is a table QA dataset sourced from statistical reports and Wikipedia pages on hierarchical tables.
\textbf{TabMWP (TabM)} \citep{lu2022dynamic} is an open-domain grade-level table question-answering dataset involving mathematical reasoning.
\textbf{TATQA (TAT)} \citep{zhu-etal-2021-tat} is a table QA dataset sourced from real-world financial reports.
\textbf{WikiTQ (Wiki)} \citep{pasupat-liang-2015-compositional} is a table QA dataset sourced from Wikipedia.
\textbf{TabFact (TabF)} \citep{Chen2020TabFact} is a table fact verification dataset sourced from Wikipedia.
\textbf{InfoTabs (Inf)} \citep{gupta-etal-2020-infotabs} is a table fact verification dataset with human-written textual hypotheses based on tables extracted from Wikipedia info-boxes.
\textbf{ToTTo (ToT)} \citep{parikh-etal-2020-totto} is a table-to-text dataset sourced from Wikipedia tables.

\paragraph{Metrics.}

For FeTaQA, we use the BLEU4 score following \citet{nan-etal-2022-fetaqa}.
For ToTTo, we follow \citet{xie-etal-2022-unifiedskg} to report the BLEU4 scores over multiple references.
We adopt the evaluation script from the original HiTab, TabMWP, TATQA, and WikiTQ repository on GitHub.
For these table QA tasks, we notice that since the fine-tuned models may not follow instructions such as ``generate in the JSON format'', we do not pose any constraints to these models in terms of the generation format.
Instead, we use Haiku 3.5\footnote{\url{https://www.anthropic.com/claude/haiku}} to extract the answer entity from the model generation. 
For TabFact and InfoTabs, we report the accuracy by checking if only the gold answer appears in the prediction.

\paragraph{Data Format.}
In terms of the test set format, we use the exact same test set for FeTaQA, HiTab, TATQA, and ToTTo as \citet{zhang-etal-2024-tablellama} with the Markdown table format.
For TabMWP, WikiTQ, and InfoTabs, etc., we follow the original data format.
Specifically, TabMWP uses `$|$' to separate columns, and WikiTQ and InfoTabs use HTML format to represent tables.

\subsection{Synthetic Table Understanding Datasets.}
In addition, we evaluate these models on eight synthesized datasets including \textbf{Beer}, \textbf{DeepM}, \textbf{Spreadsheet-DI (DI)}, \textbf{Spreadsheet-Real (ED)}, \textbf{Column-No-Separator (C)}, \textbf{Spreadsheet-CF (CF)}, and \textbf{Efthymiou (CTA)} \citep{li2023table} on schema reasoning ability (detailed in our replication for Table-GPT \Cref{subsec: mistral-tablegpt}), and \textbf{TabB\textsubscript{eval}} \citep{wu2025tablebench} on miscellaneous table tasks.

\Cref{app-sec: dataset-examples} provides examples for these datasets.

\section{Replicating Existing Table LLMs}
\label{app-sec: reimpl-compare}

\Cref{tab: tablellm-comparison} outlines the base models used in existing table LLMs. 
These base models, ranging from various Llama models to closed-source models such as GPT-3.5, differ significantly in their architecture designs, model sizes, and training recipes.
In addition, each table LLM introduces its own unique training data, making it challenging to disentangle the impact of the training data from that of the base model.
Here we report the performance of our fine-tuned models based on Mistral v0.3 7B Instruct, OLMo 7B Instruct, and Phi 3 Small Instruct (7B) versus the original models on the datasets reported in each of the original works.

\subsection{Replicating TableLlama}
\label{subsec: reimpl-tablellama}

\begin{table*}[t]
    \centering
    \small
    \renewcommand{\arraystretch}{1.3}
    \resizebox{\linewidth}{!}{
    \begin{tabular}{c|ccc|ccccccc}
    \toprule
     \multicolumn{1}{c|}{\multirow{2}{*}{Base Models}} & FeTaQA  &  HiTab & TabFact & FEVEROUS  &  HybridQA & KVRET & ToTTo & WikiSQL & WikiTQ \\
    & (BLEU) & (Acc) & (Acc) & (Acc) & (Acc) & (F1\textsubscript{Micro}) & (BLEU) & (Acc) & (Acc)\\
    \midrule
  \multicolumn{10}{l}{\grayc \textit{Original \citep{zhang-etal-2024-tablellama}}}\\
  LongLoRA 7B$^\ddagger$ & \textbf{39.0} & 64.7 & 82.5 & 73.8 & \textbf{39.4} & \underline{48.7} & 20.8 & 50.5 & 35.0 \\
\midrule
\multicolumn{10}{l}{\grayc \textit{Ours}}\\
\multicolumn{1}{c|}{Mistral v0.3 7B Instruct}  & \underline{38.7} & \textbf{70.6}$^\dagger$ & \textbf{86.8} & \underline{75.9} & 27.2 & 46.6 & \underline{28.5} & \textbf{64.5} & \underline{47.4}\\
\multicolumn{1}{c|}{OLMo 7B Instruct} & 36.8 & \underline{67.9} & 83.8 & 69.8 & 20.3 & 44.6 & 20.8 & 56.9 & 38.8 \\
\multicolumn{1}{c|}{Phi 3 Small Instruct (7B)} & 38.1 & 63.6 & \underline{86.2} & \textbf{78.3} & \underline{33.6} & \textbf{56.0} & \textbf{29.6} & \underline{63.3} & \textbf{47.7} \\

    \bottomrule
    \end{tabular}
    }
    
    \caption{Performance comparison between the original TableLlama and our fine-tuned models from different model families on the in-domain tuned (left three columns) and out-of-domain (right six columns) datasets.
    The number is bold if it is the best among the four, and underscored if it is the second.
    $\dagger$: we surpass the previous SOTA performance (64.7 by TableLlama) on HiTab.
    }
    \label{tab:mistral-tablellama}
\end{table*}

\paragraph{Training Datasets.} 
The original TableLlama \citep{zhang-etal-2024-tablellama} uses 2 million data points in its instruction tuning stage, which can be unnecessarily large.
In addition, we do not have enough computing resources to instruction-tune our model on a dataset of such a scale.
Therefore, we rule out the table operation datasets and only maintain the training data for FeTaQA \citep{nan-etal-2022-fetaqa}, HiTab \citep{cheng-etal-2022-hitab}, and TabFact \citep{Chen2020TabFact} to fine-tune our model, which results in 107K training instances.

\paragraph{Evaluation Datasets.}
Following \citet{zhang-etal-2024-tablellama}, we use the FeTaQA \citep{nan-etal-2022-fetaqa}, HiTab \citep{cheng-etal-2022-hitab}, and TabFact \citep{Chen2020TabFact} as the in-domain evaluation sets.
In addition, we compare our fine-tuned models versus the original TableLlama on FEVEROUS \citep{aly-etal-2021-fact}, HybridQA \citep{chen-etal-2020-hybridqa}, KVRET \citep{eric2017key}, ToTTo \citep{parikh-etal-2020-totto}, WikiSQL \citep{zhong2018seqsql}, and WikiTQ \citep{pasupat-liang-2015-compositional}.

\paragraph{Comparison.}
\Cref{tab:mistral-tablellama} compares the original TableLlama model (first row) versus our fine-tuned models.
Our fine-tuned models yield similar or better performance than the original TableLlama model in most cases.
In addition, we achieve the new SOTA performance on HiTab by fine-tuning the Mistral model.
As we only use 107K (5\% of the 2M data points used by the original TableLlama), our results demonstrate that \textit{with proper instruction-tuning, we can achieve competitive results on table tasks with much fewer data.}

\subsection{Replicating TableLLM}

\begin{table}[t]
    \centering
    \small
    \renewcommand{\arraystretch}{1.3}
    \setlength\tabcolsep{2pt}
    \scalebox{1}{
    \begin{tabular}{c|ccc|c}
    \toprule
     Base &  WikiTQ\textsubscript{m} & TATQA\textsubscript{m} & FeTaQA\textsubscript{m} & OTT-QA\textsubscript{m} \\
    Models & (Acc\textsubscript{p}) & (Acc\textsubscript{p}) & (BLEU) & (Acc\textsubscript{p}) \\
    \midrule
      \multicolumn{5}{l}{\grayc \textit{Original \citep{zhang2024tablellm}}}\\
    CodeLlama$^\ddagger$ & 72.5 & 51.1 & 8.4 & 57.3 \\
\midrule
\multicolumn{5}{l}{\grayc \textit{Ours}}\\
\multicolumn{1}{c|}{Mistral} & \textbf{76.0} & \underline{55.4} & \underline{10.6} & \textbf{64.3}\\
\multicolumn{1}{c|}{OLMo} & 66.8 & 50.2 & 10.5 & 58.1 \\
\multicolumn{1}{c|}{Phi} & \underline{75.4} & \textbf{57.8} & \textbf{12.1} & \underline{63.3} \\
    \bottomrule
    \end{tabular}
    }
    \caption{Performance comparison between the original TableLLM and our fine-tuned models.
    All four models are 7B and instruction-tuned.
    We denote the evaluation datasets with a subscript ``m'' as they are adapted by \citet{zhang2024tablellm}.
    }
    \label{tab:mistral-tablellm}
\end{table}

\paragraph{Training Datasets.} 
We use the original instruction-tuning set by \citet{zhang2024tablellm}, which includes 80.5K training instances.

\paragraph{Evaluation Datasets.}
Following \citet{zhang2024tablellm}, we use the modified version of WikiTQ \citep{pasupat-liang-2015-compositional}, TATQA \citep{zhu-etal-2021-tat}, and FeTaQA \citep{nan-etal-2022-fetaqa} as the in-domain evaluation sets, and OTT-QA \citep{chen2020open} as the out-of-domain evaluation set.

\paragraph{Comparison.}
\Cref{tab:mistral-tablellm} compares the original TableLLM versus our fine-tuned models.
We note that our evaluation metrics are distinct from what \citet{zhang2024tablellm} have used originally.
\citet{zhang2024tablellm} use CritiqueLLM \citep{ke-etal-2024-critiquellm} as a judge to decide the correctness of the answers.
However, the model judgments are made in Chinese\footnote{\citet{zhang2024tablellm}'s inference results are available at \url{https://github.com/RUCKBReasoning/TableLLM/blob/main/inference/results/TableLLM-7b/Grade\_fetaqa.jsonl}}, a different language from the language in all the training and evaluation datasets.
In addition, the scores assigned by the CritiqueLLM is not consistent for a single evaluation example.
Therefore, for WikiTQ\textsubscript{m}, TATQA\textsubscript{m}, and OTT-QA\textsubscript{m}, we report the Acc\textsubscript{p} scores, where we calculate whether the gold answer entities appear in the model's response.
We find that our fine-tuned models based on the Mistral and Phi models consistently outperform the original TableLLM model on these datasets, and we attribute the performance improvement to the stronger base model (Mistral v0.3 7B Instruct and Phi 3 Small Instruct) we have versus theirs (CodeLlama 7B Instruct).

\subsection{Replicating TableBenchLLM}

\begin{table}[t]
    \centering
    \small
    \renewcommand{\arraystretch}{1.3}
    \setlength\tabcolsep{2pt}
    \scalebox{1}{
    \begin{tabular}{c|cc}
    \toprule
    Base & TableBench\textsubscript{eval} \\
    Models & (R-L) \\
     \midrule
    \multicolumn{2}{l}{\grayc \textit{Original \citep{wu2025tablebench}}}\\
   Llama 3.1 8B $^\ddagger$ & \underline{27.2} \\ 
    \midrule
    \multicolumn{2}{l}{\grayc \textit{Ours}}\\
    \multicolumn{1}{c|}{Mistral v0.3 7B Instruct} & \underline{27.2} \\
    \multicolumn{1}{c|}{OLMo 7B Instruct} & 19.3 \\
    \multicolumn{1}{c|}{Phi 3 Small Instruct (7B)} & \textbf{27.8} \\
    \bottomrule
    \end{tabular}
    }
    \caption{Performance comparison between the original TablebBenchLLM based on Llama 3.1 8B and our fine-tuned models.
    ``R-L'' denotes the ROUGE-L score.
    }
    \label{tab:mistral-tablebenchllm}
\end{table}


\paragraph{Training Datasets.}
We use the original instruction-tuning set by \citet{wu2025tablebench}, which includes 20K training instances.

\paragraph{Evaluation Datasets.}
Following \citet{wu2025tablebench}, we only evaluate the model on their constructed test set, which we denote as TableBench\textsubscript{eval} in \Cref{tab:mistral-tablebenchllm}.

\paragraph{Comparison.}
Following \citet{wu2025tablebench}, we report the ROUGE-L score of our Mistral-TableBenchLLM.
In \Cref{tab:mistral-tablebenchllm}, we compare our model with the scores reported by \citet{wu2025tablebench} in the original paper, corresponding to the version of TableBenchLLM fine-tuned based on Llama 3.1 8B model.
Our Mistral-TableBenchLLM and Phi-TableBenchLLM achieve similar performance scores of 27.2 and 27.8, respectively, compared to the original TableBenchLLM's 27.2.

  \begin{table}[t]
    \centering
\small
    \renewcommand{\arraystretch}{1.3}
    \setlength\tabcolsep{2pt}
    \resizebox{\linewidth}{!}{
    \begin{tabular}{c|cccc|cccc}
    \toprule
   Base & Beer & DeepM & DI  & ED & C & CF & Wiki & CTA\\
   Models & (F1) & (Recall) & (Acc) & (F1) & (F1) & (Acc) & (Acc) & (F1)\\
    \midrule
    \multicolumn{9}{l}{\grayc \textit{Original \citep{li2023table}}} \\
    GPT-3.5$^\ddagger$ & 72.7 & \textbf{100.0} & \textbf{55.8} & \textbf{56.5} & 29.4 & \textbf{71.3} & \textbf{48.6} & \textbf{88.6} \\ 
\midrule
\multicolumn{9}{l}{\grayc \textit{Ours}} \\
\multicolumn{1}{c|}{Mistral} & \textbf{100.0} & 98.0 & 46.4 & 46.0 & 23.8 & 25.3 & 25.5 & \underline{68.3} \\
\multicolumn{1}{c|}{OLMo} & 96.2 & \textbf{100.0} & 45.4 & 35.3 & 19.9 & 29.3 & 16.4 & 62.5\\
\multicolumn{1}{c|}{Phi} & \underline{98.9} & 98.8 & \underline{49.4} & \underline{55.4} & \underline{24.8} & \underline{45.2} & \underline{30.0} & 68.3 \\
    \bottomrule
    \end{tabular}
    }
    \caption{Performance comparison between the original Table-GPT and our fine-tuned models.
    }
    \label{tab:mistral-tablegpt}
    \vspace{-0.5em}
  \end{table}%
  
  \begin{table}[t]
\centering
    \small
    \renewcommand{\arraystretch}{1.3}
    \setlength\tabcolsep{2pt}
    \scalebox{1}{
    \begin{tabular}{c|cccc|cccc}
    \toprule
   & Beer & DeepM & DI  & ED & C & CF & Wiki & CTA \\
   & (F1) & (Recall) & (Acc) & (F1) & (F1) & (Acc) & (Acc) & (F1)\\
   \midrule
    13K & 98.9 & 92.9 & 45.9 & 43.8 & \textbf{29.4} & 21.2 & 29.2 & 66.8\\
66K & \textbf{100.0} & \textbf{98.0} & \textbf{46.4} & \textbf{46.0} & 23.8 & \textbf{25.3} & \textbf{29.8} & \textbf{68.3} \\
    \bottomrule
    \end{tabular}
    }
    \caption{Performance comparison between training Mistral v0.3 7B Instruct on 13K instances versus 66K instances provided by \citet{li2023table}.
    }
    \label{tab:mistral-tablegpt-side-finding-all}
  \end{table}

\subsection{Replicating Table-GPT}
\label{subsec: mistral-tablegpt}

\paragraph{Training Dataset.}
We use the instruction-tuning dataset provided by \citet{li2023table} that contains 66K instances.

\paragraph{Evaluation Datasets.}
We select four in-domain test sets by \citet{li2023table}, Beer for entity matching, DeepM for schema matching, Spreadsheet-DI (DI) for data imputation, and Spreadsheet-Real (ED) for error detection.
Furthermore, we report the out-of-domain performance on Column-No-Separator (C) for missing value identification, Spreadsheet-CF (CF) for column finding, WikiTQ (Wiki) for table question answering, and Efthymiou (CTA) for column type annotation.

\paragraph{Comparison.}

\Cref{tab:mistral-tablegpt} reports the results.
We note that though the size of our fine-tuned models are all 7B, they achieve better performance than Table-GPT which is based on GPT-3.5 on Beer, and comparable performance on DeepM.
However, on the out-of-domain datasets, we can see that Mistral-TableGPT underperforms the original Table-GPT.
We attribute such performance differences to the differences between the base models.
Since GPT-3.5 is stronger than these open-source 7B models, its innate table understanding ability, as well as its generalization ability, leads to better performance on these out-of-domain table datasets for Table-GPT.
This reinforces our motivations of conducting the comparisons using the same base model, as \textit{the performance difference may be because of the base model's capability}, therefore, we need the same base model to conduct an apple-to-apple comparison.


\paragraph{Side Findings.}
There is a smaller training set provided by \citet{li2023table} containing 13K training instances.
We report the performance comparison by training the Mistral v0.3 7B Instruct model on the two sets in \Cref{tab:mistral-tablegpt-side-finding-all}
We do not find a significant performance boost when we use the larger 66K dataset.
And on one of the out-of-domain datasets, C, training on 13K instances even yields a better score of 29.4 than training on 66K instances' 23.8.
This echoes with the findings by \citet{zhou2024lima,deng2025rethinking} that limited instruction tuning instances are able to yield a strong model.

\section{Results and Discussions}
\label{app-sec: results-and-discussions}


\begin{figure}[!t]
    \centering
    \begin{subfigure}[t]{0.35\textwidth}
    \includegraphics[width=0.95\linewidth]{imgs/correlations/shapley_legend.crop.pdf}
    \end{subfigure}
    \vspace{0.3em}
    
    \begin{subfigure}[t]{0.45\textwidth}
    \includegraphics[width=0.95\linewidth]{imgs/correlations/shapley_Three.crop.pdf}
    \caption{All three models considered.}
    \end{subfigure}
     \begin{subfigure}[t]{0.45\textwidth}
    \includegraphics[width=0.95\linewidth]{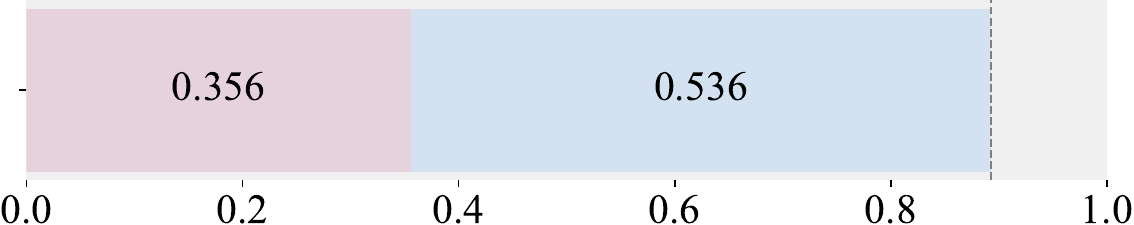}
    \caption{Mistral and OLMo.}
    \end{subfigure}
     \begin{subfigure}[t]{0.45\textwidth}
    \includegraphics[width=0.95\linewidth]{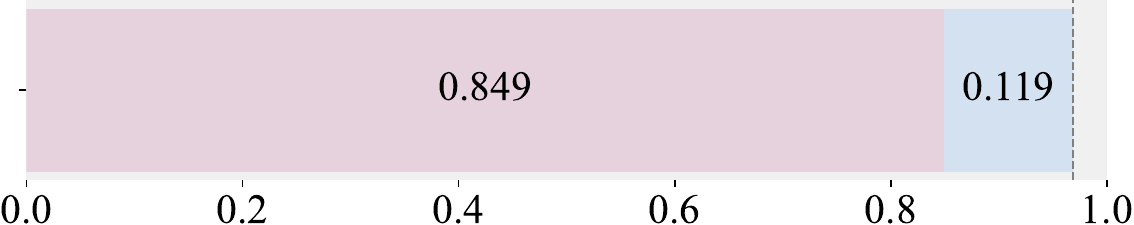}
    \caption{Mistral and Phi.}
    \end{subfigure}
     \begin{subfigure}[t]{0.45\textwidth}
    \includegraphics[width=0.95\linewidth]{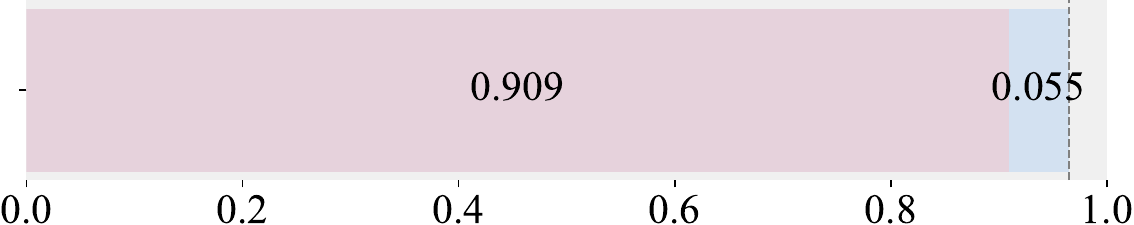}
    \caption{OLMo and Phi.}
    \end{subfigure}
    \caption{Shapley R$^2$ decomposition \citep{shapley1953value, israeli2007shapley} for the contributions of the downstream tasks' performance by the base LLM versus the training set.
    We can see that the choice of the base LLM is a non-negligible factor, and in many cases, the dominant factor that decides the model's performance on downstream tasks.
    }
    \label{fig:shapley}
\end{figure}

\subsection{Shapley R$^2$ Decomposition}
\label{app-subsec: shapley-decomp}

To quantify the relative contributions of (i) the base LLM and (ii) the instruction-tuning dataset to downstream table-understanding performance, we apply a Shapley R$^2$ decomposition. Our implementation follows the standard model-agnostic formulation for linear regression and corresponds directly to our released code.

\paragraph{Setup.}

For each of the 12 instruction-tuned models, we compute the average performance across the OOD datasets, yielding a target vector

\[
y \in \mathbb{R}^{12}.
\]

We consider two categorical predictors:
\[
\text{Base model} \in \{\text{Mistral-7B}, \text{OLMo-7B}, \text{Phi-3-7B}\},
\]

\[
\begin{aligned}
    \text{Train set} \in & \{\text{TableLlama}, \text{TableLLM}, \\
    & \text{TableBenchLLM}, \text{Table-GPT}\}.
\end{aligned}
\]

Both variables are one-hot encoded, producing the design matrix

\[
X = \big[ X_{\text{Base}},\; X_{\text{TrainSet}} \big].
\]

Shapley $R^{2}$ for Two Feature Groups.

For two predictor groups, the Shapley value for group $i$ is

\[
\begin{aligned}
    \phi_i = \frac{1}{2}\!\left( R^2(\{i\}) - R^2(\varnothing) \right)
+ \\\frac{1}{2}\!\left(R^2(\{\text{Base}, \text{TrainSet}\}) - R^2(\{j\}) \right),
\end{aligned}
\]

where $i \neq j$, and $R^2(S)$ denotes the coefficient of determination of a linear regression model using only the feature subset $S$.

Concretely, $R^2(\{\text{Base}\})$ uses only base-model indicators, $R^2(\{\text{TrainSet}\})$ uses only instruction-tuning dataset indicators, and $R^2(\{\text{Base},\text{TrainSet}\})$ is the full model $R^2$.

\paragraph{Permutation-Based Formulation.}

More generally, our implementation follows the full permutation-based Shapley formulation:

\[
\phi_i
= \frac{1}{|\Pi|}
\sum_{\pi \in \Pi}
\left(
R^{2}\big( S_i^\pi \cup \{i\} \big)
- R^{2}\big( S_i^\pi \big)
\right),
\]

where $\Pi$ is the set of all permutations of \{\text{Base}, \text{TrainSet}\}, and $S_i^\pi$ is the set of feature groups that appear before $i$ in permutation $\pi$.
With two feature groups, there are exactly two permutations (Base → TrainSet and TrainSet → Base), and we average marginal contributions across both.

\paragraph{Implementation Details.}

We compute $R^{2}$ using ordinary least squares via \texttt{sklearn}'s \texttt{LinearRegression}. Categorical predictors are one-hot encoded, and Shapley values are accumulated across permutations and normalized such that

\[
\phi_{\text{Base}} + \phi_{\text{TrainSet}} = R^2_{\text{Full Model}}.
\]

\paragraph{Interpretation.}

The Shapley values quantify how much variance in table-understanding performance is attributable to (i) base model identity versus (ii) the instruction-tuning dataset. Empirically, we obtain:
\[
\phi_{\text{Base}} = 0.816, \qquad
\phi_{\text{TrainSet}} = 0.138.
\]

These results indicate that base model choice overwhelmingly dominates the contribution of instruction-tuning data in determining the performance of table-tuned LLMs.

\paragraph{Results and Discussions.}
\Cref{fig:shapley} provides the Shapley R$^2$ results for the three models as well as for each pair of models.
We note that when we consider model pairs, base model selection is a dominant factor that decides the instruction-tuned models' performance for Mistral and Phi, OLMo and Phi.
For models fine-tuned from Mistral and OLMo, base model selection still explains 35.6\% of the performance variance. 
This suggests that the base model selection is a crucial, and in many cases, a dominant factor that determines the instruction-tuned model's performance.




\begin{figure*}[t]
    \centering
    \begin{subfigure}[t]{0.48\linewidth}
        \includegraphics[width=0.95\linewidth]{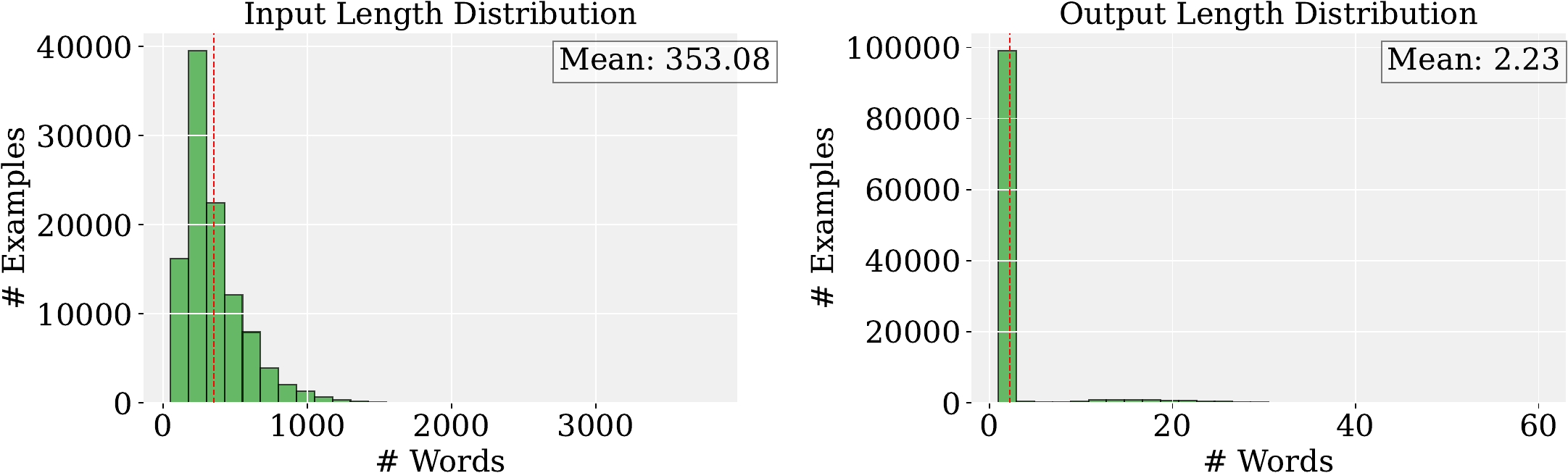}
        \caption{TableLLlama.}
    \end{subfigure}
    \begin{subfigure}[t]{0.48\linewidth}
        \includegraphics[width=0.95\linewidth]{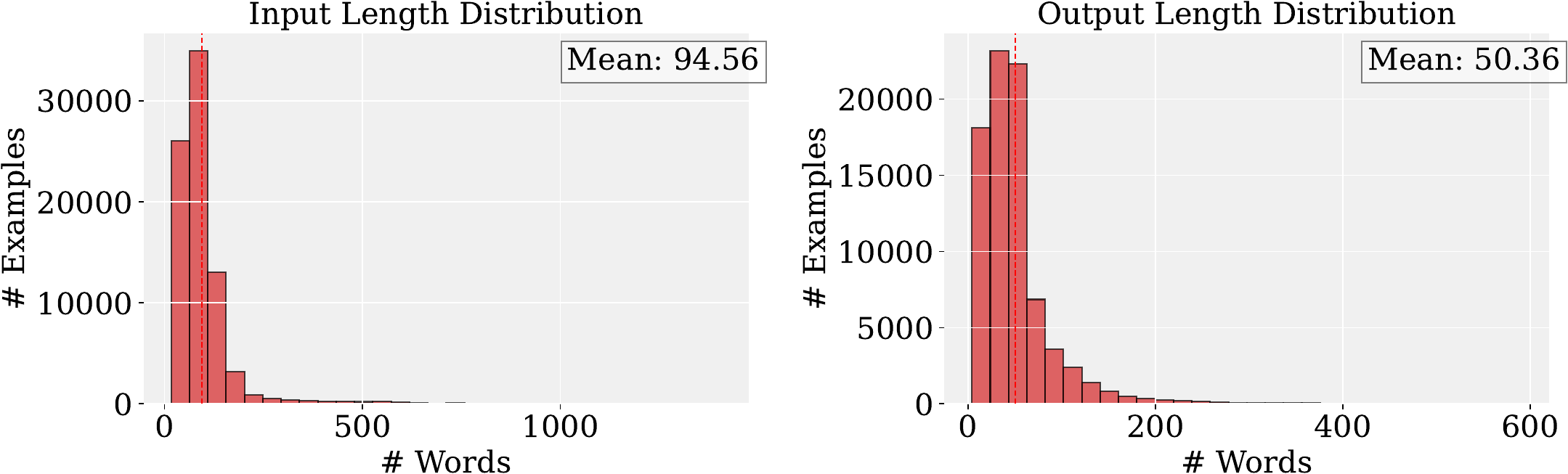}
        \caption{TableLLM.}
    \end{subfigure}
    \begin{subfigure}[t]{0.48\linewidth}
        \includegraphics[width=0.95\linewidth]{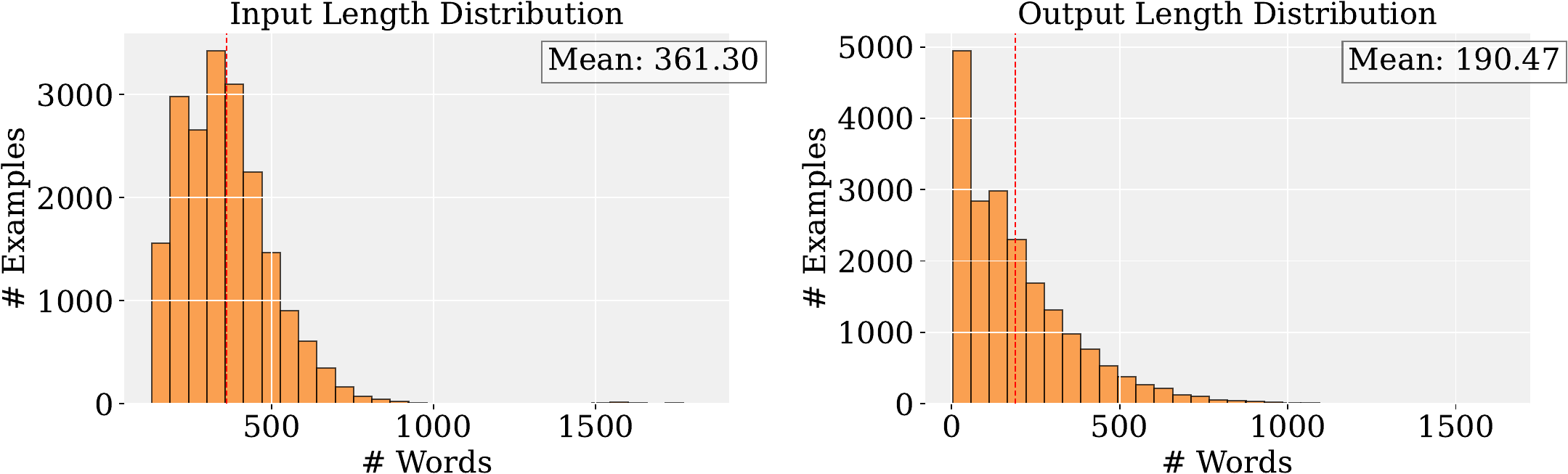}
        \caption{TableBench.}
        \label{subfig: tablebench-distr}
    \end{subfigure}
    \begin{subfigure}[t]{0.48\linewidth}
        \includegraphics[width=0.95\linewidth]{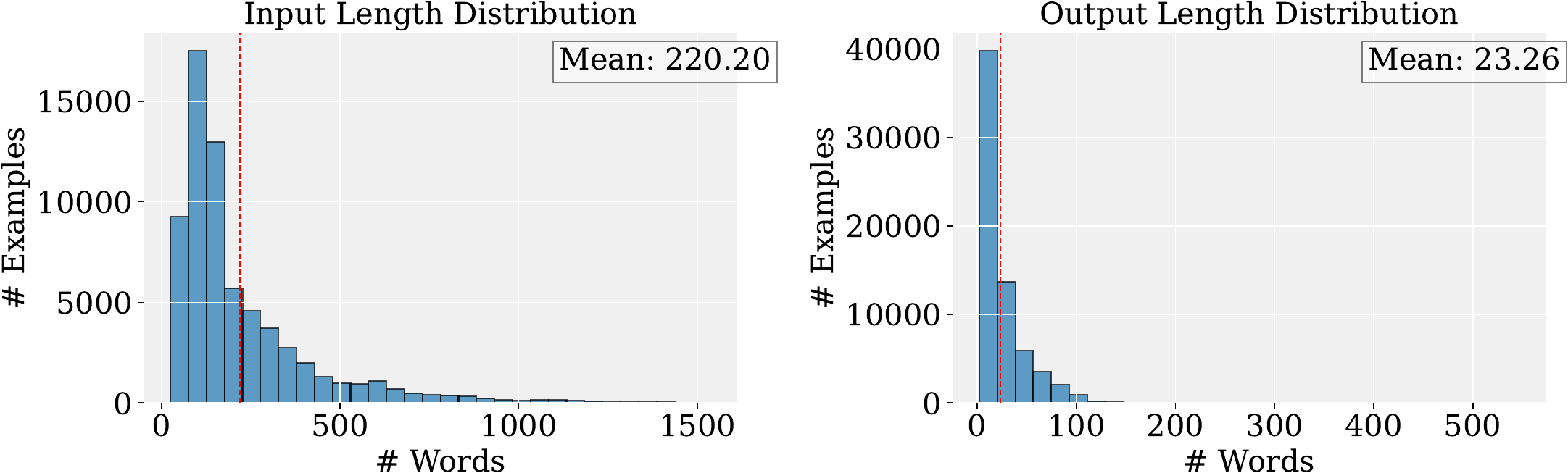}
        \caption{TableGPT.}
    \end{subfigure}
    \caption{Distributions of the training data in terms of the input length and output length.}
    \label{fig:hist-train-data}
\end{figure*}

\begin{table*}[t]
        \small
    \centering
    \renewcommand{\arraystretch}{1.3}
    \setlength\tabcolsep{5pt}
    \scalebox{1}{
    \begin{tabular}{cp{14cm}}
        \toprule
        \multicolumn{2}{l}{\grayc \textit{TableLlama \citep{zhang-etal-2024-tablellama}}}\\
        Q &  What was the percent of restaurants and bars that were frequently the setting of behaviours related to unwanted physical contact or suggested sexual relations that happened off campus? \\
        \graymidrule
        Gold & 49.4.\\
        \midrule
        \multicolumn{2}{l}{\grayc \textit{TableLLM \citep{zhang2024tablellm}}}\\
        Q &  How many works did Leyla Erbil publish in total? \\
        \graymidrule
        \multirow{2}{*}{Gold} &  Leyla Erbil published a total of 11 works. 
        \textit{This can be determined by counting the number of entries in the ``Name'' column in the provided table}.\\
        \midrule
        \multicolumn{2}{l}{\grayc \textit{TableBench \citep{wu2025tablebench}}}\\
        \multirow{2}{*}{Q} &  Can you compare the performance of the advocates based on their wins, losses, and poll results, and identify which advocate has the most balanced performance across all metrics? \\
        \graymidrule
        \multirow{3}{*}{Gold} & The table lists various advocates along with their performance metrics in terms of wins, losses, ties, poll wins, and poll losses. Patton Oswalt shows the most balanced performance across all metrics with 3 wins, 2 losses, 1 tie, 3 poll wins, and 3 poll losses.\\
        \midrule
        \multicolumn{2}{l}{\grayc \textit{TableGPT \citep{li2023table}}}\\
        Q &  predict the output value for the last row denoted as '[Output Value].'  \\
        \graymidrule
        Gold &  6406 m.\\
        \bottomrule
    \end{tabular}
    }
    \captionof{table}{Training examples from TableLlama, TableLLM, TableBench, and TableGPT.
    We omit the corresponding table here for readability.
    The reasoning part is in italics for TableLLM data.
    }
    \label{tab:tablellm-example}
\end{table*}

\subsection{Training Data Example}
\label{appx-subsec: train-data-examples}

As shown in \Cref{tab:tablellm-example}, the training instance from TableLLM contains the underlying reasoning process to reach the final answer.
Such traces would benefit the model's reasoning process, as suggested by the findings by \citet{guo2025deepseek, muennighoff2025s1}.
\Cref{fig:hist-train-data} displays the distributions of input and output lengths across training datasets. 
Notably, TableLlama exhibits significantly shorter output lengths compared to other training datasets. 
While TableBench has the longest average output length, its distribution possesses a high frequency of single-word answers (the prominent peak in the output distribution in \Cref{subfig: tablebench-distr}). 
Furthermore, TableBench outputs may contain irrelevant reasoning elements (the first half of the gold answer is not relevant to the comparison of the performance in \Cref{tab:tablellm-example}).

\subsection{RQ5: How does the table instruction tuning compromise the general capabilities of the foundation LLMs?}
\label{app-subsec: affect-general-capabilities}

\paragraph{Evaluation Setup.}
We select five general benchmarks.
\textbf{MMLU} \citep{hendryckstest2021} examines the general ability of the model on 57 tasks including elementary mathematics, US history, computer science, etc.
We adopt the 5-shot setup.
\textbf{MMLU\textsubscript{Pro}} \citep{wang2024mmlupro} is an enhanced benchmark evaluating the general ability of the model, which contains up to ten options and eliminates the trivial questions in MMLU.
We adopt the 5-shot setup.
\textbf{AI2ARC} \citep{clark2018think} is a reasoning benchmark containing natural, grade-school questions. 
We adopt the 0-shot setup and report the accuracy score on the challenging set.
\textbf{GPQA} \citep{rein2023gpqa} is a reasoning benchmark containing questions in biology, physics, and chemistry written by domain experts. 
We adopt a 0-shot setup and report the accuracy score on its main set.
\textbf{IFEval} \citep{zhou2023instruction} is a dataset evaluating the general instruction following ability of the model containing instructions such as ``return the answer in JSON format''.
We report the instance-level strict accuracy defined by \citet{zhou2023instruction}.
We include examples from these datasets in \Cref{app-sec: dataset-examples}.

For MMLU, MMLU\textsubscript{Pro}, AI2ARC, and GPQA, as they are all multi-choice question-answering datasets, our objective is to select the most appropriate completion among a set of given options based on the provided context. 
Following \citet{touvron2023llama}, we select the completion with the highest likelihood given the provided context. 
As we evaluate the model based on their selection of the letter choice of ``A'', ``B'', etc., we do not normalize the likelihood by the number of characters in the completion.

\paragraph{Answer: Table instruction tuning does not necessarily compromise the base models' general capabilities.}
\Cref{fig:general-benchmark-performance-difference} provides the model's performance on the five general benchmarks, while \Cref{tab:general} provides the performance in numbers.
We find that on MMLU, MMLU\textsubscript{Pro}, AI2ARC, and GPQA, \textit{our fine-tuned models do not compromise too much of the base models' general capabilities.}
On AI2ARC, the score for Mistral-TableGPT is even slightly higher than the base model.
Such performance improvement is likely due to the fact that many table tasks involve reasoning over tables, which may enhance the model's general reasoning ability.
On IFEval, models fine-tuned from the Mistral model suffer a significant performance drop of over 20 points compared to the original model.
However, models fine-tuned from the Phi model even improve the base model's performance.
Contrary to the works arguing that tuning would compromise the model's capabilities \citep{luo2023empirical}, our finding suggests that domain-specific tuning does not necessarily lead to performance decay on general benchmarks, and the base model selection plays a crucial role in maintaining base LLMs' general capabilities.


    
    


\subsection{RQ6: How does the model size affect performance on table tasks?}
\label{app-subsec: model-size}

\paragraph{Evaluation Setup.}
We compare Phi 3 Mini Instruct (4B) versus Phi 3 Small Instruct (7B) on the table benchmarks introduced in \Cref{app-sec: eval-benchmarks}.

\begin{figure*}[t]
    \centering
    \begin{subfigure}[t]{0.3\textwidth}
        \centering
        \includegraphics[width=\linewidth]{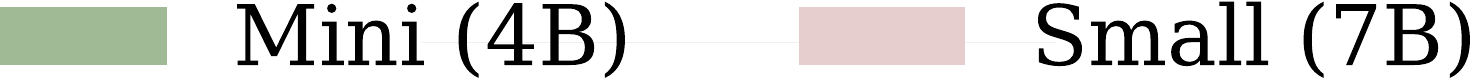}
    \end{subfigure}%
    
    \vspace*{1em}
     \begin{subfigure}[t]{0.95\textwidth}
        \centering
        \includegraphics[width=\linewidth]{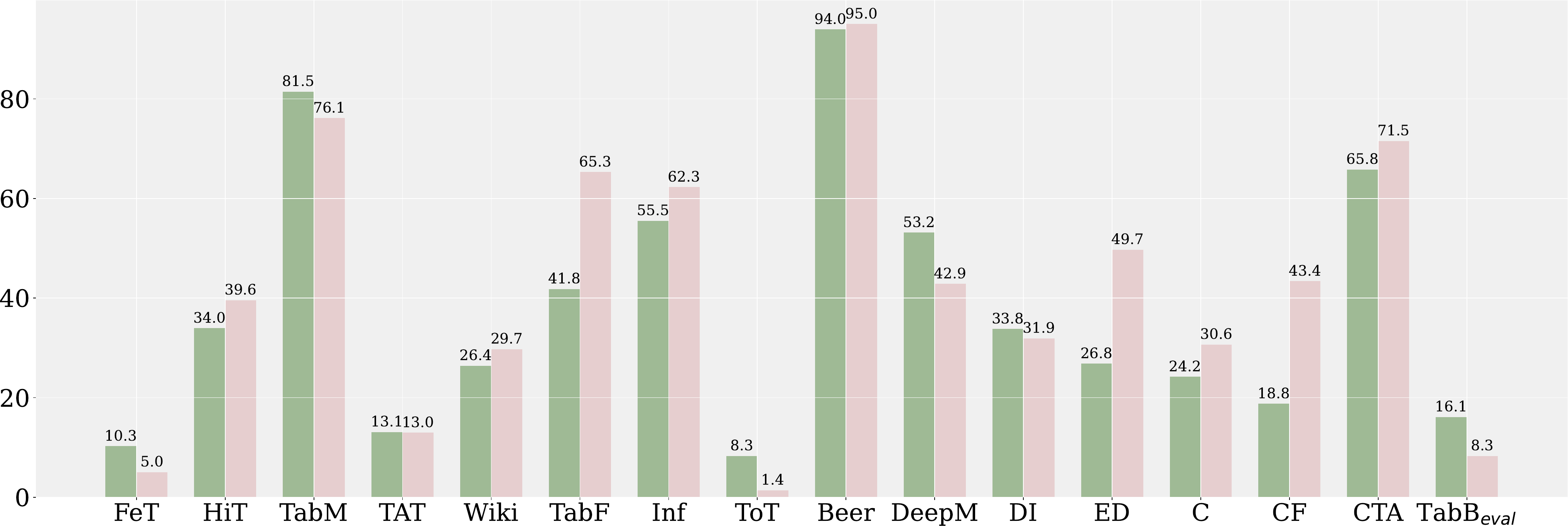}
        \caption{No training data, the original model.}
    \end{subfigure}%
    \vspace*{1em}
     \begin{subfigure}[t]{0.95\textwidth}
        \centering
        \includegraphics[width=\linewidth]{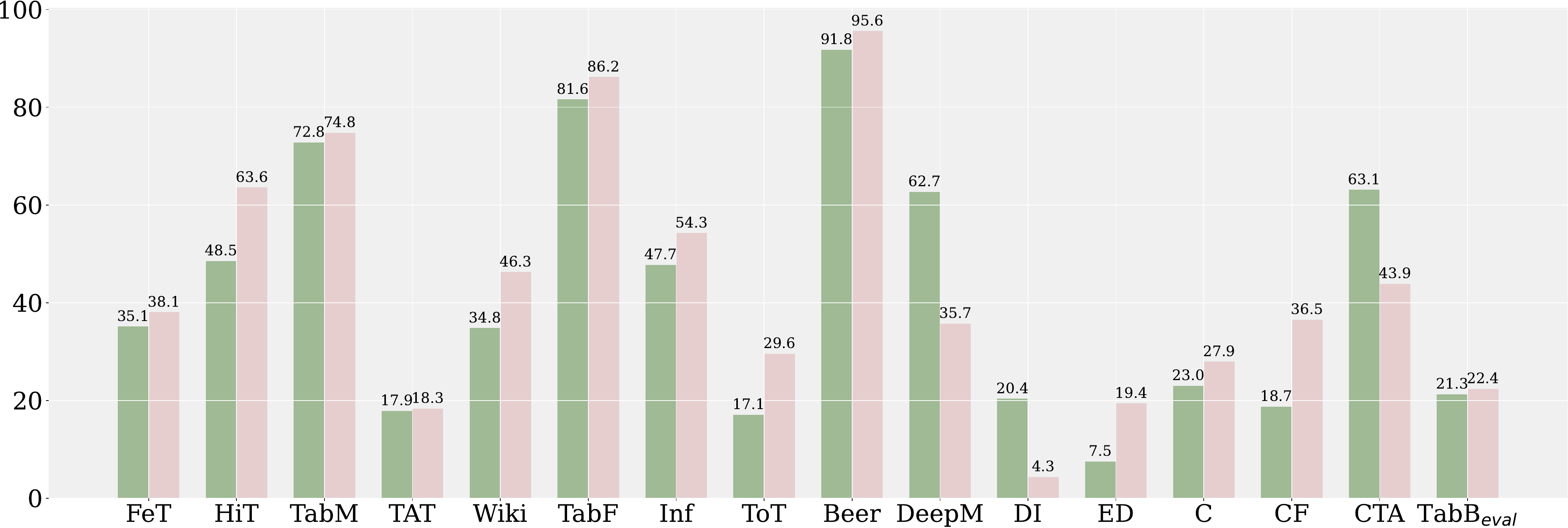}
        \caption{Training data for TableLlama.}
    \end{subfigure}%
    \vspace*{1em}
     \begin{subfigure}[t]{0.95\textwidth}
        \centering
        \includegraphics[width=\linewidth]{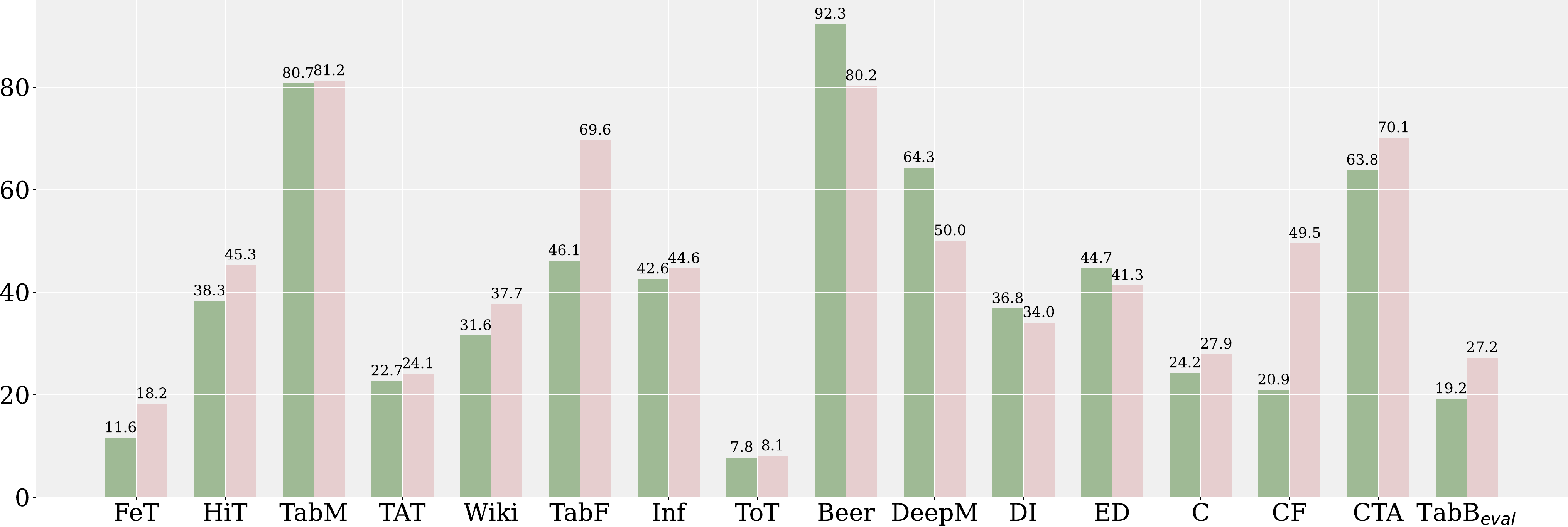}
        \caption{Training data for TableLLM.}
    \end{subfigure}%

    \caption{Performance of Phi 3 Mini Instruct (4B) versus Phi 3 Small Instruct (7B) model on different table tasks with different training data.
    In most cases, the 7B model outperforms the 4B model.
    }
    \label{fig:table-benchmark-size-1}
    
\end{figure*}

\begin{figure*}[t]
    \centering
    \begin{subfigure}[t]{0.3\textwidth}
        \centering
        \includegraphics[width=\linewidth]{imgs/bar_general_diff_size/size_comparison_general_legend.crop.pdf}
    \end{subfigure}%
    
    \vspace*{1em}
     \begin{subfigure}[t]{0.95\textwidth}
        \centering
        \includegraphics[width=\linewidth]{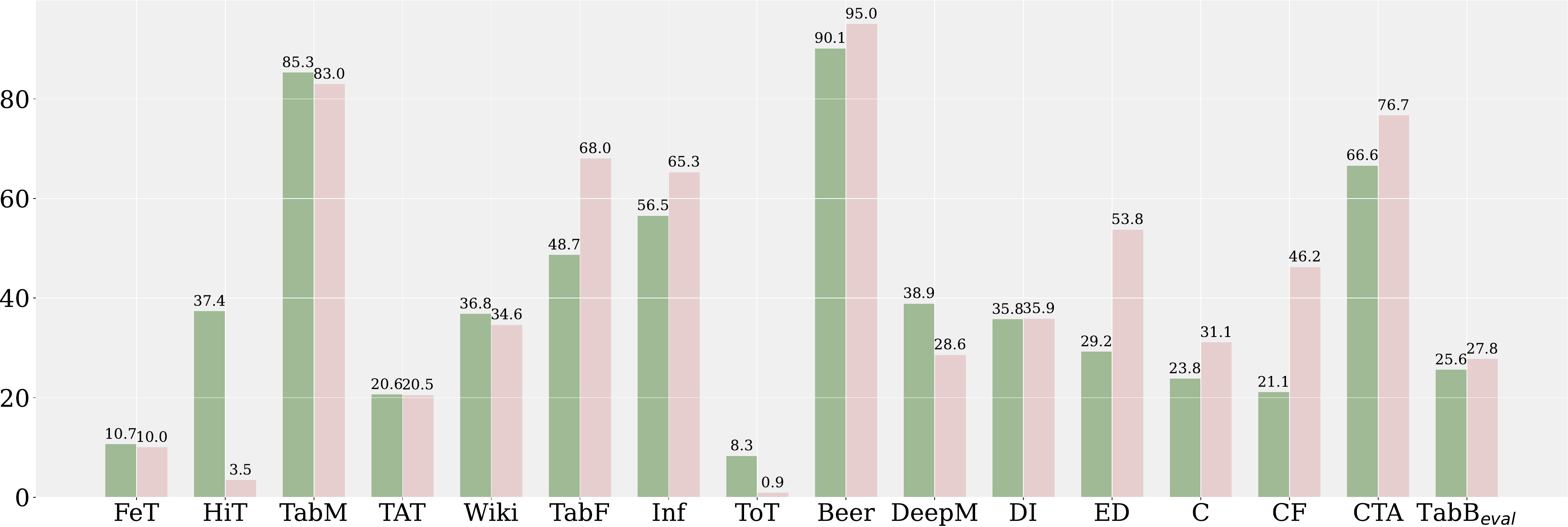}
        \caption{Training data for TableBench.}
    \end{subfigure}%
    \vspace*{1em}
     \begin{subfigure}[t]{0.95\textwidth}
        \centering
        \includegraphics[width=\linewidth]{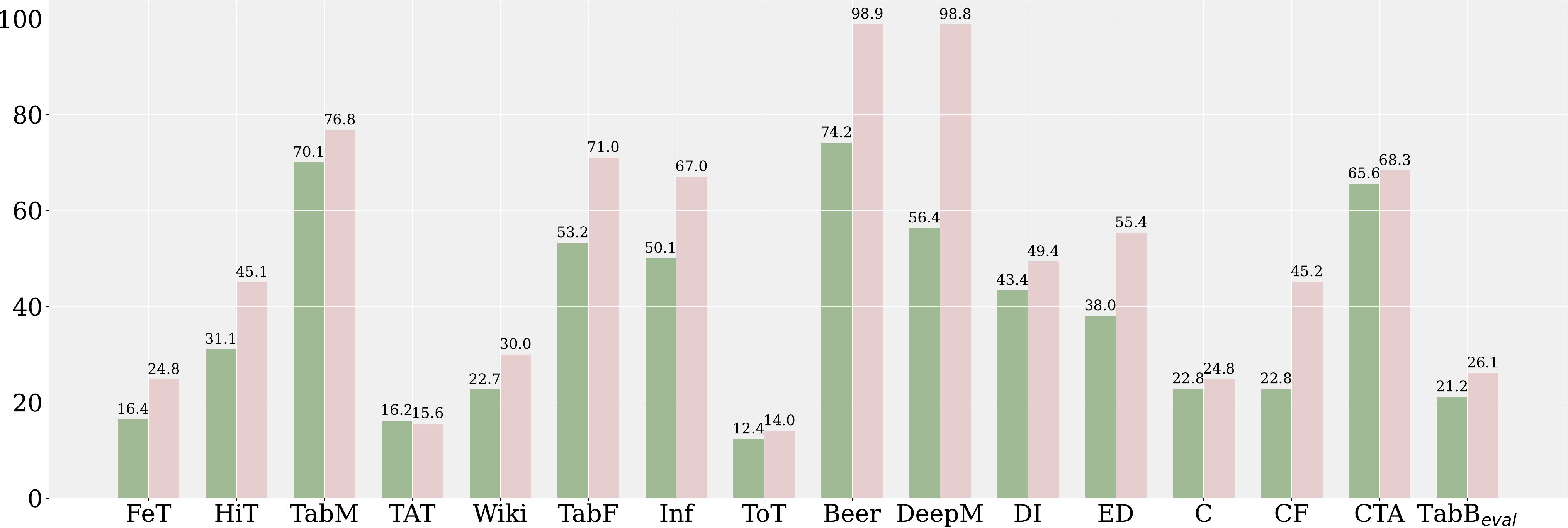}
        \caption{Training data for TableGPT.}
    \end{subfigure}%

    \caption{Performance of Phi 3 Mini Instruct (4B) versus Phi 3 Small Instruct (7B) model on different table tasks with different training data.
    In most cases, the 7B model outperforms the 4B model.
    }
    \label{fig:table-benchmark-size-2}
    
\end{figure*}

\paragraph{Answer: Model performance increases as the model size becomes larger.}
\Cref{fig:table-benchmark-size-1,fig:table-benchmark-size-2} provide a performance comparison between Phi 3 Mini Instruct (4B) versus Phi 3 Small Instruct (7B).
Similar to the findings for the general LLMs \citep{dubey2024llama, wei2022emergent}, we find that the larger-sized model often leads to better performance for both the original model and the model after training on the same set of data.

Additionally, we present \textbf{the trade-offs between GPU hours and the model performance} in \Cref{tab: gpu-hours-tradeoff}.
These results demonstrate a consistent trend: the 7B model outperforms the 4B variant in all settings. However, the improvements are modest (e.g., +2.24 points in the TableLlama setting) relative to the increase in computational cost ($\sim$35\% more GPU hours).

This highlights the trade-off: while larger models can yield stronger performance, the marginal gains may be insufficient to justify the additional cost in resource-constrained environments.

\begin{table*}[t]
\small
\centering
\begin{tabular}{l
>{\columncolor[HTML]{FFFDFA}}l 
>{\columncolor[HTML]{FFFDFA}}l 
>{\columncolor[HTML]{FFFDFA}}l 
>{\columncolor[HTML]{FFFDFA}}l }
\toprule
\textbf{Training Data} & {\color[HTML]{333333} \textbf{Base Model}}  & {\color[HTML]{333333} \textbf{Model Sizes}} & {\color[HTML]{333333} \textbf{GPU Hours}} & {\color[HTML]{333333} \textbf{Avg Performance}} \\
\midrule
N/A                    & {\color[HTML]{333333} Phi 3 Mini Instruct}  & {\color[HTML]{333333} 4B}                   & {\color[HTML]{333333} N/A}                & {\color[HTML]{333333} 39.19}                    \\
N/A                    & {\color[HTML]{333333} Phi 3 Small Instruct} & {\color[HTML]{333333} 7B}                   & {\color[HTML]{333333} N/A}                & {\color[HTML]{333333} 41.42}                    \\
TableLlama             & {\color[HTML]{333333} Phi 3 Mini Instruct}  & {\color[HTML]{333333} 4B}                   & {\color[HTML]{333333} 441.2}              & {\color[HTML]{333333} 39.25}                    \\
TableLlama             & {\color[HTML]{333333} Phi 3 Small Instruct} & {\color[HTML]{333333} 7B}                   & {\color[HTML]{333333} 596.0}                & {\color[HTML]{333333} 41.49}                    \\
TableLLM               & {\color[HTML]{333333} Phi 3 Mini Instruct}  & {\color[HTML]{333333} 4B}                   & {\color[HTML]{333333} 333.2}              & {\color[HTML]{333333} 42.38}                    \\
TableLLM               & {\color[HTML]{333333} Phi 3 Small Instruct} & {\color[HTML]{333333} 7B}                   & {\color[HTML]{333333} 442.8}              & {\color[HTML]{333333} 42.41}    \\
\bottomrule
\end{tabular}
\caption{Training GPU hours versus the averaged model performance for the Phi models at different sizes.
We average the model performance across the OOD table tasks.
These results demonstrate a consistent trend: the 7B model outperforms the 4B variant in all settings. However, the improvements are modest (e.g., +2.24 points in the TableLlama setting) relative to the increase in computational cost ($\sim$35\% more GPU hours).
}
\label{tab: gpu-hours-tradeoff}
\end{table*}

\section{Additional Discussions}
\label{appx-sec: additional-discussion}
\subsection{Future Directions}
\label{app-subsec: future-directions}

\paragraph{Toward better table benchmarks.}
As LLMs continue to advance rapidly \citep{ouyang2022training, touvron2023llama, dubey2024llama, yang2024qwen2}, there is a growing need for a comprehensive evaluation of table-related capabilities. 
Existing benchmarks often focus on narrow domains or specific subtasks \citep{Chen2020TabFact, nan-etal-2022-fetaqa}, while recent work has begun to explore broader coverage through synthetic datasets \citep{wu2025tablebench} and multi-table reasoning setups \citep{wu2025mmqa}. 
However, concerns remain regarding the gap between synthetic benchmarks and authentic user needs. 
Future work shall ground table benchmarks in real-world use cases and build datasets that more accurately reflect user-driven queries and interactions with structured data.

\paragraph{Incorporating prior insights from table modeling.}
In the era of table LLMs, most efforts have focused on instruction tuning and dataset construction \citep{zhang-etal-2024-tablellama, zheng-etal-2024-multimodal}. 
Yet, earlier work in table modeling demonstrates that incorporating table-specific features and structure-aware model architectures can significantly improve performance \citep{herzig-etal-2020-tapas, yang-etal-2022-tableformer}.
We advocate for future research to revisit and integrate these insights into modern table modeling, potentially bridging architecture-level innovations with instruction tuning strategies.

One promising direction is to integrate table-specific inductive biases—originally explored in models like TableFormer \citep{yang-etal-2022-tableformer} and TAPAS \citep{herzig-etal-2020-tapas}—into the instruction-tuning pipeline of modern LLMs such as Phi or Mistral.
For instance, table-aware attention mechanisms that distinguish between row-wise and column-wise relationships \citep{yang-etal-2022-tableformer} can be incorporated during fine-tuning. 
These architectural modifications can guide the model to better preserve structural information by restricting attention across table rows and columns, thereby encoding relational dependencies more explicitly.

While modifying the core architecture of pretrained models like Phi or Mistral may not be practical, such inductive biases can be \textit{approximated} during instruction tuning by using

\begin{enumerate}
    \item Input-level encoding schemes (e.g., special tokens or segment IDs to mark rows and columns) \citep{herzig-etal-2020-tapas}.
    \item Adapter-based approaches where additional lightweight modules are trained to encode structural priors \citep{hu2023llm}.
    \item Reparameterize attention patterns via learned masks or prompts that encourage the model to attend differently across rows and columns—simulating table-aware attention without altering the model architecture \citep{lester2021power, li2021prefix}.
\end{enumerate}

\paragraph{Bridging techniques from other fields.}
Table modeling has a long-standing tradition of adapting techniques from other areas of NLP \citep{yin-etal-2020-tabert}. 
Recent efforts leverage vision-language models \citep{deng-etal-2024-tables, zheng-etal-2024-multimodal}.
In this paper, we endeavor to leverage meta-evaluation \citep{kobayashi2024revisiting, veuthey2025meqa} to scrutinize the existing table evaluation framework. 
Here we list two future directions: (1) employing mechanistic interpretability methods \citep{huben2024sparse} to better understand how models represent and reason over structured inputs; and (2) leveraging membership inference attacks \citep{shokri2017membership} to probe the potential leakage or memorization of structured data in pretraining corpora.

\paragraph{Bringing structures to the broader NLP.}
While table modeling often borrows from other subfields, we believe that table research can benefit the broader NLP community.
\citet{hawkins2021thousand} suggest that inherent structures\footnote{\citet{hawkins2021thousand} refer to these structures as ``reference frame''.} exist in human reasoning, and recent works suggest that LLMs can benefit from reasoning with structures \citep{sun2025table}.
Reasoning in structures can potentially lead to more robust, interpretable, and modularized output \citep{wang2024chainoftable}.
We encourage future efforts on this and potentially bringing insights into the table research to the broader NLP community.

\subsection{Comparing Against Proprietary LLMs}
\label{app-subsec: compare-propriety-llms}

Due to the immense computational cost and lack of public fine-tuning APIs (e.g., for Claude or GPT-4), our study —like most prior table LLM work \citep{zhang-etal-2024-tablellama} — focuses on 7B-scale open-source models such as Mistral, OLMo, and Phi. 
Despite these constraints, we attempt to contextualize our results by benchmarking our instruction-tuned 7B models against available larger or commercial models on public datasets.
\Cref{tab: tabfact-acc-propriety,tab: wiki-acc-overall} are some of the comparisons (TabFact, WikiTableQuestions) we are able to compile.

In \Cref{tab: tabfact-acc-propriety}, our instruction-tuned models match or even outperform commercial models like GPT-4o. 
For instance, Phi (7B) trained with TableLlama reaches 86.2 on TabFact, compared to GPT-4o’s reported 84.1.

In \Cref{tab: wiki-acc-overall}, while GPT-4 performs strongly on Wiki, some 7B-scale instruction-tuned models—such as Phi trained on TableLlama—demonstrate competitive performance.

For practitioners and researchers (e.g., start-ups or regulated domains) who lack access to proprietary APIs or massive compute budgets, our findings provide a replicable, useful snapshot of what is possible. 
For instruction tuning at the 7B scale, our findings explain why some small-model recipes succeed while others fail.

\begin{table}[t]
\small
\resizebox{\linewidth}{!}{
\begin{tabular}{
>{\columncolor[HTML]{FFFDFA}}l 
>{\columncolor[HTML]{FFFDFA}}l 
>{\columncolor[HTML]{FFFDFA}}l 
>{\columncolor[HTML]{FFFDFA}}l 
>{\columncolor[HTML]{FFFDFA}}l 
>{\columncolor[HTML]{FFFDFA}}l }
\toprule
{\color[HTML]{333333} \textbf{Base Model}} & {\color[HTML]{333333} \textbf{Size}} & {\color[HTML]{333333} \textbf{\begin{tabular}[c]{@{}l@{}}Training \\ Data\end{tabular}}} & {\color[HTML]{333333} \textbf{ID?}} & {\color[HTML]{333333} \textbf{\begin{tabular}[c]{@{}l@{}}Reasoning \\ Model?\end{tabular}}} & {\color[HTML]{333333} \textbf{Acc}}  \\
\midrule
{\color[HTML]{333333} o4-mini}             & {\color[HTML]{333333} Unk}           & {\color[HTML]{333333} N/A}                                                               & {\color[HTML]{333333} No}           & {\color[HTML]{333333} Yes}                                                                  & {\color[HTML]{2C3A4A} \textbf{94.3}} \\
{\color[HTML]{333333} Deepseek-R1}         & {\color[HTML]{333333} 685B}          & {\color[HTML]{333333} N/A}                                                               & {\color[HTML]{333333} No}           & {\color[HTML]{333333} Yes}                                                                  & {\color[HTML]{333333} 92.2}          \\
{\color[HTML]{333333} GPT-3.5}             & {\color[HTML]{333333} Unk}           & {\color[HTML]{333333} N/A}                                                               & {\color[HTML]{333333} No}           & {\color[HTML]{333333} No}                                                                   & {\color[HTML]{333333} 67.4}         \\
{\color[HTML]{333333} GPT-4}               & {\color[HTML]{333333} Unk}           & {\color[HTML]{333333} N/A}                                                               & {\color[HTML]{333333} No}           & {\color[HTML]{333333} No}                                                                   & {\color[HTML]{333333} 74.4}          \\
{\color[HTML]{333333} GPT-4o}              & {\color[HTML]{333333} Unk}           & {\color[HTML]{333333} N/A}                                                               & {\color[HTML]{333333} No}           & {\color[HTML]{333333} No}                                                                   & {\color[HTML]{2C3A4A} \textbf{84.1}} \\
{\color[HTML]{333333} Llama 3}             & {\color[HTML]{333333} 70B}           & {\color[HTML]{333333} N/A}                                                               & {\color[HTML]{333333} No}           & {\color[HTML]{333333} No}                                                                   & {\color[HTML]{2C3A4A} 79.4} \\
{\color[HTML]{333333} Llama 3}             & {\color[HTML]{333333} 8B}            & {\color[HTML]{333333} N/A}                                                               & {\color[HTML]{333333} No}           & {\color[HTML]{333333} No}                                                                   & {\color[HTML]{333333} 58.4}         \\
{\color[HTML]{333333} Mistral}             & {\color[HTML]{333333} 7B}            & {\color[HTML]{333333} N/A}                                                               & {\color[HTML]{333333} No}           & {\color[HTML]{333333} No}                                                                   & {\color[HTML]{333333} 62.3}          \\
{\color[HTML]{333333} OLMo}                & {\color[HTML]{333333} 7B}            & {\color[HTML]{333333} N/A}                                                               & {\color[HTML]{333333} No}           & {\color[HTML]{333333} No}                                                                   & {\color[HTML]{333333} 38.2}          \\
{\color[HTML]{333333} Phi}                 & {\color[HTML]{333333} 7B}            & {\color[HTML]{333333} N/A}                                                               & {\color[HTML]{333333} No}           & {\color[HTML]{333333} No}                                                                   & {\color[HTML]{2C3A4A} 65.3} \\
{\color[HTML]{333333} Phi-Mini}            & {\color[HTML]{333333} 4B}            & {\color[HTML]{333333} N/A}                                                               & {\color[HTML]{333333} No}           & {\color[HTML]{333333} No}                                                                   & {\color[HTML]{333333} 41.8}          \\
{\color[HTML]{333333} Mistral}             & {\color[HTML]{333333} 7B}            & {\color[HTML]{333333} Table-GPT}                                                         & {\color[HTML]{333333} No}           & {\color[HTML]{333333} No}                                                                   & {\color[HTML]{333333} 61.4}          \\
{\color[HTML]{333333} OLMo}                & {\color[HTML]{333333} 7B}            & {\color[HTML]{333333} Table-GPT}                                                         & {\color[HTML]{333333} No}           & {\color[HTML]{333333} No}                                                                   & {\color[HTML]{333333} 44.9}          \\
{\color[HTML]{333333} Phi}                 & {\color[HTML]{333333} 7B}            & {\color[HTML]{333333} Table-GPT}                                                         & {\color[HTML]{333333} No}           & {\color[HTML]{333333} No}                                                                   & {\color[HTML]{333333} 71.0}            \\
{\color[HTML]{333333} Phi-Mini}            & {\color[HTML]{333333} 4B}            & {\color[HTML]{333333} Table-GPT}                                                         & {\color[HTML]{333333} No}           & {\color[HTML]{333333} No}                                                                   & {\color[HTML]{333333} 53.2}          \\
{\color[HTML]{333333} Mistral}             & {\color[HTML]{333333} 7B}            & {\color[HTML]{333333} TableLlama}                                                        & {\color[HTML]{333333} Yes}          & {\color[HTML]{333333} No}                                                                   & {\color[HTML]{2C3A4A} \textbf{86.8}} \\
{\color[HTML]{333333} OLMo}                & {\color[HTML]{333333} 7B}            & {\color[HTML]{333333} TableLlama}                                                        & {\color[HTML]{333333} Yes}          & {\color[HTML]{333333} No}                                                                   & {\color[HTML]{333333} 83.8}          \\
{\color[HTML]{333333} Phi}                 & {\color[HTML]{333333} 7B}            & {\color[HTML]{333333} TableLlama}                                                        & {\color[HTML]{333333} Yes}          & {\color[HTML]{333333} No}                                                                   & {\color[HTML]{333333} 86.2}          \\
{\color[HTML]{333333} Phi-Mini}            & {\color[HTML]{333333} 4B}            & {\color[HTML]{333333} TableLlama}                                                        & {\color[HTML]{333333} Yes}          & {\color[HTML]{333333} No}                                                                   & {\color[HTML]{333333} 81.6}     \\    
\bottomrule
\end{tabular}}
\caption{Model performance on TabFact dataset. 
Our instruction-tuned models match or even outperform commercial models like GPT-4o. For instance, Phi (7B) trained with TableLlama reaches 86.2 on TabFact, compared to GPT-4o’s 84.1.
}
\label{tab: tabfact-acc-propriety}
\end{table}

\begin{table}[t]
\small
\centering
\begin{tabular}{
>{\columncolor[HTML]{FFFDFA}}l 
>{\columncolor[HTML]{FFFDFA}}l 
>{\columncolor[HTML]{FFFDFA}}l 
>{\columncolor[HTML]{FFFDFA}}l }
\toprule
{\color[HTML]{333333} \textbf{Base Model}} & {\color[HTML]{333333} \textbf{Model Size}} & {\color[HTML]{333333} \textbf{\begin{tabular}[c]{@{}l@{}}Training \\ Data\end{tabular}}} & {\color[HTML]{333333} \textbf{Acc}}  \\
\midrule
{\color[HTML]{333333} GPT-3.5}             & {\color[HTML]{333333} Unk}                 & {\color[HTML]{333333} N/A}                                                               & {\color[HTML]{2C3A4A} \textbf{20.9}} \\
{\color[HTML]{333333} GPT-4}               & {\color[HTML]{333333} Unk}                 & {\color[HTML]{333333} N/A}                                                               & {\color[HTML]{2C3A4A} \textbf{55.9}} \\
{\color[HTML]{333333} Mistral}             & {\color[HTML]{333333} 7B}                  & {\color[HTML]{333333} N/A}                                                               & {\color[HTML]{333333} 27.9}          \\
{\color[HTML]{333333} OLMo}                & {\color[HTML]{333333} 7B}                  & {\color[HTML]{333333} N/A}                                                               & {\color[HTML]{333333} 19.4}          \\
{\color[HTML]{333333} Phi}                 & {\color[HTML]{333333} 7B}                  & {\color[HTML]{333333} N/A}                                                               & {\color[HTML]{333333} 29.7}          \\
{\color[HTML]{333333} Phi-Mini}            & {\color[HTML]{333333} 7B}                  & {\color[HTML]{333333} N/A}                                                               & {\color[HTML]{333333} 26.4}          \\
{\color[HTML]{333333} GPT-3.5}             & {\color[HTML]{333333} Unk}                 & {\color[HTML]{333333} Table-GPT}                                                         & {\color[HTML]{333333} 48.6}          \\
{\color[HTML]{333333} Mistral}             & {\color[HTML]{333333} 7B}                  & {\color[HTML]{333333} Table-GPT}                                                         & {\color[HTML]{333333} 25.5}          \\
{\color[HTML]{333333} OLMo}                & {\color[HTML]{333333} 7B}                  & {\color[HTML]{333333} Table-GPT}                                                         & {\color[HTML]{333333} 16.4}          \\
{\color[HTML]{333333} Phi}                 & {\color[HTML]{333333} 7B}                  & {\color[HTML]{333333} Table-GPT}                                                         & {\color[HTML]{333333} 30.0}            \\
{\color[HTML]{333333} Phi-Mini}            & {\color[HTML]{333333} 4B}                  & {\color[HTML]{333333} Table-GPT}                                                         & {\color[HTML]{333333} 22.7}          \\
{\color[HTML]{333333} Mistral}             & {\color[HTML]{333333} 7B}                  & {\color[HTML]{333333} TableLlama}                                                        & {\color[HTML]{333333} 23.8}          \\
{\color[HTML]{333333} OLMo}                & {\color[HTML]{333333} 7B}                  & {\color[HTML]{333333} TableLlama}                                                        & {\color[HTML]{333333} 6.7}           \\
{\color[HTML]{333333} Phi}                 & {\color[HTML]{333333} 7B}                  & {\color[HTML]{333333} TableLlama}                                                        & {\color[HTML]{2C3A4A} \textbf{46.3}} \\
{\color[HTML]{333333} Phi-Mini}            & {\color[HTML]{333333} 4B}                  & {\color[HTML]{333333} TableLlama}                                                        & {\color[HTML]{333333} 34.8}   \\
\bottomrule
\end{tabular}
\caption{
Model performance on Wiki. 
While GPT-4 performs strongly on Wiki, some 7B-scale instruction-tuned models—such as Phi trained on TableLlama—demonstrate competitive performance.}
\label{tab: wiki-acc-overall}
\end{table}

\subsection{Additional Datasets}
\label{app-subec: addi-datasets}

One concern we received is that the 16 benchmarks may not fully represent the scale and messiness of real-world enterprise tables.

To address such a concern, we conduct additional evaluation on six tasks on table QA and real-world table operations, including data integration, semantic enrichment, and schema-level reasoning \citep{xing2025mmtu, chen2021finqa, cafarella2008webtables, he2015sema, abedjan2016dataxformer} listed in \Cref{tab: addi-data}.

\begin{table}[t]
\small
\centering
\begin{tabular}{p{5em}p{12em}c}
\toprule
\textbf{Categories}   & \textbf{Task Description}                    & \textbf{Metrics} \\
\midrule
Table-QA              & Answer questions based on tables             & Acc              \\
Data-Imputation       & Predict missing values in tables             & Acc              \\
Semantic-transform    & Predict semantic transformations by examples & Acc              \\
Entity-Matching       & Match rows refer to the same semantic entity & Acc              \\
Functional-Dependency & Predict functional-dependency in tables      & F1               \\
Semantic-join         & Predict semantic join between two tables     & Acc         \\
\bottomrule
\end{tabular}
\caption{Additional datasets for testing.}
\label{tab: addi-data}
\end{table}

We report the performance of several 7B models (including our instruction-tuned variants) in \Cref{tab: addi-results}.
Our experiments reveal substantial performance variance across models, underscoring the central finding of our paper: base model selection plays a critical role. 
In particular, Phi 3 Small consistently outperforms OLMo across these tasks when trained with comparable instruction data (e.g. TableLLM’s training data).

\begin{table*}[t]
\small
\centering
\begin{tabular}{llllllll}
\toprule
\textbf{Dataset} & \textbf{\begin{tabular}[c]{@{}l@{}}OLMo 7B \\ Instruct\end{tabular}} & \textbf{\begin{tabular}[c]{@{}l@{}}OLMo 7B \\ Instruct\end{tabular}} & \textbf{\begin{tabular}[c]{@{}l@{}}Phi 3 \\ Small\end{tabular}} & \textbf{\begin{tabular}[c]{@{}l@{}}Phi 3 \\ Small\end{tabular}} & \textbf{\begin{tabular}[c]{@{}l@{}}Phi 3 \\ Small\end{tabular}} & \textbf{\begin{tabular}[c]{@{}l@{}}Phi 3 \\ Small\end{tabular}} & \textbf{\begin{tabular}[c]{@{}l@{}}Phi 3 \\ Small\end{tabular}} \\
\midrule
Size                 & 7B                                                                   & 7B                                                                   & 7B                                                              & 7B                                                              & 7B                                                              & 7B                                                              & 7B                                                              \\
   Train Data              &                                           N/A                           & TableLLM                                                             & N/A                                                             & TableLlama                                                      & TableLLM                                                        & TableBench                                                      & TableGPT                                                        \\
   \midrule
FinQA            & 0.7                                                                  & 0.1                                                                  & 7.2                                                             & \textbf{20.0}                                                   & 6.8                                                             & 7.2                                                             & 10.7                                                            \\
WebTable         & 0.2                                                                  & 4.3                                                                  & 10.5                                                            & 9.1                                                             & 14.3                                                            & 15.9                                                            & 30.1                                                            \\
SEMA-Join        & 2.0                                                                  & 17.5                                                                 & 69.4                                                            & 72.2                                                            & 67.8                                                            & \textbf{72.9}                                                   & 70.6                                                            \\
Fodors-Zagats    & 49.5                                                                 & 67.2                                                                 & 97.8                                                            & 61.8                                                            & \textbf{99.5}                                                   & 98.9                                                            & \textbf{99.5}                                                   \\
DBLP-Scholar     & 77.4                                                                 & 86.2                                                                 & 93.7                                                            & 43.0                                                            & \textbf{95.6}                                                   & 94.4                                                            & 95.3                                                            \\
Auto-Relate      & 0.0                                                                  & 0.0                                                                  & \textbf{39.5}                                                   & 3.0                                                             & 36.5                                                            & 28.4                                                            & 24.2                                                            \\
DataXFormer      & 0.0                                                                  & 0.0                                                                  & 39.2                                                            & 31.8                                                            & \textbf{47.1}                                                   & 42.0                                                            & 32.7                                                            \\
\midrule
AVG              & 18.5                                                                 & 25.0                                                                 & 51.1                                                            & 34.4                                                            & \textbf{52.5}                                                   & 51.4                                                            & 51.9          \\
\bottomrule
\end{tabular}
\caption{Model performance on additional datasets listed in \Cref{tab: addi-data}. 
Our experiments reveal substantial performance variance across models, underscoring the central finding of our paper: base model selection plays a critical role. 
In particular, Phi 3 Small consistently outperforms OLMo across these tasks when trained with comparable instruction data (e.g. TableLLM’s training data).
}
\label{tab: addi-results}
\end{table*}

\input{figures/general-benchmark-performance-difference}

\begin{table}[t]
    \small
    \centering
    \renewcommand{\arraystretch}{1.3}
    \setlength\tabcolsep{2pt}
    \resizebox{\linewidth}{!}{
    \begin{tabular}{cccccc}
    \hline
        \multicolumn{1}{c|}{\multirow{2}{*}{\textbf{Method}}} & \multicolumn{1}{c|}{MMLU} &  \multicolumn{1}{c|}{MMLU\textsubscript{Pro}} & \multicolumn{1}{c|}{AI2ARC} & \multicolumn{1}{c|}{GPQA} & \multicolumn{1}{c}{IFEval} \\
        \cline{2-6}
        & \multicolumn{1}{|c|}{Acc} & \multicolumn{1}{c|}{Acc} & \multicolumn{1}{c|}{Acc} & \multicolumn{1}{c|}{Acc} & \multicolumn{1}{c}{Acc}\\
        \hline
M    & 61.2 & 31.4 & 73.3 & 28.6 & 58.8 \\
M-TableLlama & 59.4 & 29.5 & 69.6 & 23.7 & 38.0 \\
$\Delta$ & \hspace{-0.2em}$\downarrow$ 1.9 & \hspace{-0.2em}$\downarrow$ 1.9 & \hspace{-0.2em}$\downarrow$ 3.4 & \hspace{-0.2em}$\downarrow$ 4.9 & \hspace{-0.3em}$\downarrow$ 20.7 \\
\midrule
M-TableLLM & 61.4 & 29.3 & 74.2 & 25.9 & 29.6 \\
$\Delta$ & \hspace{-0.2em}$\uparrow$ 0.2 & \hspace{-0.2em}$\downarrow$ 2.0 & \hspace{-0.2em}$\uparrow$ 0.9 & \hspace{-0.2em}$\downarrow$ 2.7 & \hspace{-0.3em}$\downarrow$ 29.1 \\
\midrule
M-TableBenchLLM & 62.0 & 31.0 & 73.6 & 28.1 & 31.8 \\
$\Delta$ & \hspace{-0.2em}$\uparrow$ 0.7 & \hspace{-0.2em}$\downarrow$ 0.4 & \hspace{-0.2em}$\uparrow$ 0.3 & \hspace{-0.2em}$\downarrow$ 0.5 & \hspace{-0.3em}$\downarrow$ 27.0 \\
\midrule
M-TableGPT & 61.3 & 31.3 & 74.6 & 26.1 & 31.4 \\
$\Delta$ & \hspace{-0.2em}$\uparrow$ 0.1 & \hspace{-0.2em}$\downarrow$ 0.1 & \hspace{-0.2em}$\uparrow$ 1.3 & \hspace{-0.2em}$\downarrow$ 2.4 & \hspace{-0.3em}$\downarrow$ 27.3 \\
\hline
\hline
O & 52.6 & 22.5 & 67.6 & 27.9 & 45.6 \\
O-TableLlama & 53.7 & 23.1 & 66.2 & 29.7 & 46.8 \\
$\Delta$ & $\uparrow$ 1.1 & $\uparrow$ 0.6 & $\downarrow$ 1.4 & $\uparrow$ 2.0 & $\uparrow$ 1.2 \\\hline
O-TableLLM & 53.3 & 22.3 & 66.0 & 29.0 & 42.8 \\
$\Delta$ & $\uparrow$ 0.7 & $\downarrow$ 0.3 & $\downarrow$ 1.6 & $\uparrow$ 1.9 & $\downarrow$ 2.8 \\\hline
O-TableBenchLLM & 53.1 & 21.9 & 67.7 & 28.6 & 45.2 \\
$\Delta$ & $\uparrow$ 0.5 & $\downarrow$ 0.7 & $\uparrow$ 0.1 & $\uparrow$ 0.9 & $\downarrow$ 0.4 \\\hline
O-TableGPT & 52.9 & 21.9 & 66.8 & 28.8 & 48.9 \\
$\Delta$ & $\uparrow$ 0.3 & $\downarrow$ 0.6 & $\downarrow$ 0.8 & $\uparrow$ 0.8 & $\uparrow$ 3.4 \\
\hline
\hline
P & 75.7 & 41.2 & 73.1 & 31.0 & 60.7 \\
P-TableLlama & 75.5 & 45.1 & 73.5 & 31.5 & 70.1 \\
$\Delta$ & $\downarrow$ 0.2 & $\uparrow$ 3.9 & $\uparrow$ 0.3 & $\uparrow$ 0.4 & $\uparrow$ 9.9 \\\hline
P-TableLLM & 75.0 & 42.6 & 73.1 & 30.4 & 64.8 \\
$\Delta$ & $\downarrow$ 0.7 & $\uparrow$ 1.3 & $\uparrow$ 0.0 & $\downarrow$ 0.8 & $\uparrow$ 4.1 \\\hline
P-TableBenchLLM & 75.7 & 43.3 & 60.8 & 28.8 & 63.3 \\
$\Delta$ & $\uparrow$ 0.0 & $\uparrow$ 2.0 & $\downarrow$ 1.5 & $\downarrow$ 2.1 & $\uparrow$ 2.6 \\\hline
P-TableGPT & 75.1 & 40.1 & 72.6 & 32.4 & 70.0 \\
$\Delta$ & $\downarrow$ 0.5 & $\downarrow$ 1.2 & $\downarrow$ 0.3 & $\uparrow$ 1.4 & $\uparrow$ 9.4 \\
\hline

    \end{tabular}
    }
    \caption{Evaluation of the models on general benchmarks.
    ``M-'', ``O-'', and ``P-'' represent Mistral v0.3 7B Instruct, OLMo 7B Instruct, Phi 3 Small Instruct (7B), respectively.
    ``$\Delta$'' denotes the performance difference between the instruction-tuned model and its base model.
    }
    \label{tab:general}
\end{table}

\section{Dataset Examples}
\label{app-sec: dataset-examples}

\subsection{FeTaQA}

\textbf{Input:}
\begin{lstlisting}
[TLE] The Wikipedia page title of this table is Gerhard Bigalk. The Wikipedia section title of this table is Ships attacked. [TAB] | Date | Name | Nationality | Tonnage (GRT) | Fate | [SEP] | 14 June 1941 | St. Lindsay | United Kingdom | 5,370 | Sunk | [SEP] | 21 December 1941 | HMS Audacity | Royal Navy | 11,000 | Sunk | [SEP] | 2 February 1942 | Corilla | Netherlands | 8,096 | Damaged | [SEP] | 4 February 1942 | Silveray | United Kingdom | 4,535 | Sunk | [SEP] | 7 February 1942 | Empire Sun | United Kingdom | 6,952 | Sunk | [SEP] | 16 May 1942 | Nicarao | United States | 1,445 | Sunk | [SEP] | 19 May 1942 | Isabela | United States | 3,110 | Sunk |\n\nThe highlighted cells of the table are: [HIGHLIGHTED_BEGIN] [11,000], [Sunk], [8,096], [Damaged] [HIGHLIGHTED_END] What happened to the two heaviest ships Gerhard Bigalk attacked?
\end{lstlisting}

\textbf{Instruction:}
\begin{lstlisting}
This is a free-form table question answering task. The goal for this task is to answer the given question based on the given table and the highlighted cells.
\end{lstlisting}
\textbf{Output:}
\begin{lstlisting}
Gerhard Bigalk damaged one ship of 8,096 GRT, and sunk one warship of 11,000 tons.
\end{lstlisting}

\subsection{TabFact}
\textbf{Input:}
\begin{lstlisting}
[TLE] The table caption is about tony lema. [TAB] | tournament | wins | top - 5 | top - 10 | top - 25 | events | cuts made [SEP] | masters tournament | 0 | 1 | 2 | 4 | 4 | 4 | [SEP] | us open | 0 | 2 | 3 | 4 | 6 | 5 | [SEP] | the open championship | 1 | 2 | 2 | 2 | 3 | 3 | [SEP] | pga championship | 0 | 0 | 1 | 2 | 5 | 4 | [SEP] | totals | 1 | 5 | 8 | 12 | 18 | 16 |\n\nThe statement is:  <tony lema be in the top 5 for the master tournament , the us open , and the open championship>. Is it entailed or refuted by the table above?
\end{lstlisting}
\textbf{Instruction:}
\begin{lstlisting}
This is a table fact verification task. The goal of this task is to distinguish whether the given statement is entailed or refuted by the given table.
\end{lstlisting}
\textbf{Output:}
\begin{lstlisting}
entailed
\end{lstlisting}

\subsection{ToTTo}
\textbf{Input:}
\begin{lstlisting}
<page_title> List of Governors of South Carolina </page_title> <section_title> Governors under the Constitution of 1868 </section_title> <table> <cell> 76 <col_header> # </col_header> <col_header> 74 </col_header> <col_header> 75 </col_header> </cell> <cell> Daniel Henry Chamberlain <col_header> Governor </col_header> <row_header> 76 </row_header> </cell> <cell> December 1, 1874 <col_header> Took Office </col_header> <row_header> 76 </row_header> </cell> </table>
\end{lstlisting}
\textbf{Instruction:}
\begin{lstlisting}
This is a highlighted cells description task. The goal of this task is to generate the language description given table cells.
\end{lstlisting}
\textbf{Output:}
\begin{lstlisting}
Daniel Henry Chamberlain was the 76th Governor of South Carolina from 1874.
\end{lstlisting}

\subsection{Beer}
\textbf{Input:}
\begin{lstlisting}
Beer A is:\n|name|factory|\n|---|---|\n|Sierra Amber Ale|Silver Peak Restaurant \& Brewery|\n\nBeer B is:\n|name|factory|\n|---|---|\n|Sierra Andina Alpamayo Amber Ale|Sierra Andina|    
\# Task Description: Please determine whether Beer A and Beer B refer to the same entity or not.
\end{lstlisting}
\textbf{Instruction:}
\begin{lstlisting}
You are a helpful assistant that specializes in tables.\n Your final answer should be \'Yes\' or \'No\'. Return the final result as JSON in the format \{"answer": "<Yes or No>"\}.  Let's think step by step and show your reasoning before showing the final result.
\end{lstlisting}
\textbf{Output:}
\begin{lstlisting}
\{"answer": "No"\}
\end{lstlisting}

\subsection{TabB\textsubscript{eval}}

\textbf{Input:}
\begin{lstlisting}
Read the table below in JSON format:\n[TABLE] \n\{"columns": ["index", "organization", "year", "rank", "out of"], "data": [["bribe payers index", "transparency international", 2011, 19, 28], ["corruption perceptions index", "transparency international", 2012, 37, 176], ["democracy index", "economist intelligence unit", 2010, 36, 167], ["ease of doing business index", "world bank", 2012, 16, 185], ["economic freedom index", "fraser institute", 2010, 15, 144], ["economic freedom index", "the heritage foundation", 2013, 20, 177], ["global competitiveness report", "world economic forum", 20122013, 13, 144], ["global peace index", "institute for economics and peace", 2011, 27, 153], ["globalization index", "at kearney / foreign policy magazine", 2006, 35, 62], ["press freedom index", "reporters without borders", 2013, 47, 179], ["property rights index", "property rights alliance", 2008, 28, 115]]\}\n\nLet\'s get start!\nQuestion: What is the average rank of the indices published by Transparency International?
\end{lstlisting}
\textbf{Instruction:}
\begin{lstlisting}
You are a helpful assistant that specializes in tables.\nYou are a table analyst. Your task is to answer questions based on the table content.\n\n\nThe answer should follow the format below:\n[Answer Format]\nFinal Answer: AnswerName1, AnswerName2...\n\nEnsure the final answer format is the last output line and can only be in the "Final Answer: AnswerName1, AnswerName2..." form, no other form. Ensure the "AnswerName" is a number or entity name, as short as possible, without any explanation.\n\n\nGive the final answer to the question directly without any explanation.
\end{lstlisting}
\textbf{Output:}
\begin{lstlisting}
28
\end{lstlisting}

\subsection{MMLU}
\textbf{Input:}
\begin{lstlisting}
{5-shot examples}
Find the degree for the given field extension Q(sqrt(2), sqrt(3), sqrt(18)) over Q.	
\nA. 0\nB. 4\nC. 2\nD. 6\nAnswer:
\end{lstlisting}
\textbf{Instruction:}
\begin{lstlisting}
The following are multiple choice questions (with answers) about abstract algebra.\n\n
\end{lstlisting}
\textbf{Output:}
\begin{lstlisting}
B
\end{lstlisting}

\subsection{IFEval}
\textbf{Input:}
\begin{lstlisting}
Can you help me make an advertisement for a new product? It's a diaper that's designed to be more comfortable for babies and I want the entire output in JSON format.
\end{lstlisting}
\textbf{Instruction:}
\begin{lstlisting}
You are a helpful assistant.
\end{lstlisting}
\textbf{Output:}
\begin{lstlisting}
[JSON formatted answer]
\end{lstlisting}

\end{document}

%% file: figures/paradigm-shift.tex
\begin{figure*}[!ht]
    \centering
\begin{minipage}[t]{0.45\textwidth}  
    \vspace{0pt}
    \begin{subfigure}[t]{\textwidth}
        \centering
        \includegraphics[width=\linewidth]{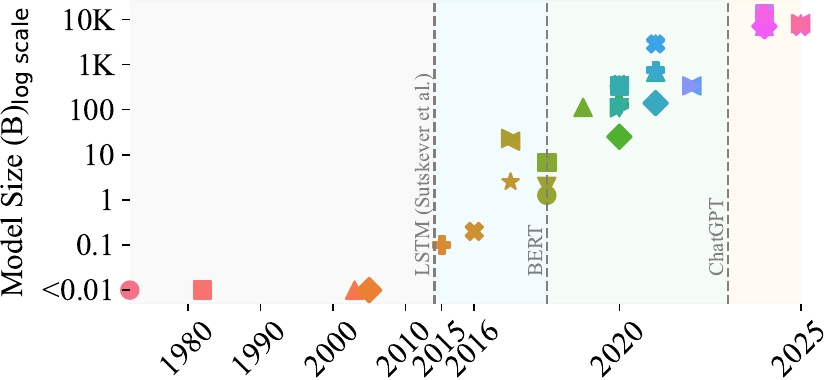}
    \end{subfigure}%
    \small
    \renewcommand{\arraystretch}{1.3}
    \begin{tabular}{l}
   \colorbox{eraRule}{\parbox{\linewidth}{\bfseries\textcolor{black}{Rule-based} \scriptsize up to 2014}}\\
   {\it Characteristics}: symbolic, static (\citeyear{woods1972lunar}) \\
   {\it Challenges}: labor-intensive; generalization (\citeyear{warren-pereira-1982-efficient})\\
    \colorbox{eraSeq2Seq}{\parbox{\linewidth}{\bfseries\textcolor{black}{Seq2Seq} \scriptsize 2014\(\sim \)2018}}\\
    \begin{tabular}{@{}l@{}}
    {\it Characteristics}: feature engineering (\citeyear{zhong2018seqsql}) \\
   {\it Challenges}: generalization (\citeyear{yu-etal-2018-spider});\\
   \hspace{5em} reasoning \citep{liu2018table}
   \end{tabular}\\
    \colorbox{eraTransformer}{\parbox{\linewidth}{\bfseries\textcolor{black}{Transformer} \scriptsize 2018\(\sim \)2023}}\\
    \begin{tabular}{@{}l@{}}
    {\it Characteristics}: domain pre-training (\citeyear{yin-etal-2020-tabert}); \\ 
    \hspace{5em}feature engineering (\citeyear{wang-etal-2020-rat})
    \end{tabular}\\
    \begin{tabular}{@{}l@{}}
   {\it Challenges}: generalization (\citeyear{suhr-etal-2020-exploring});\\
   \hspace{5em}data-intensive (\citeyear{yin-etal-2020-tabert}), etc
   \end{tabular}\\
    \colorbox{eraLLM}{\parbox{\linewidth}{\bfseries\textcolor{black}{LLM Era} \scriptsize 2023 to now}}\\
    {\it Characteristics}: post-training (\citeyear{zhang-etal-2024-tablellama, yang2025table})\\
    \begin{tabular}{@{}l@{}}
   {\it Challenges}: generalization (\citeyear{deng2025rethinking});\\
   \hspace{5em}paradox of choice (\citeyear{zhang-etal-2024-tablellama, zhang2024tablellm}), etc \\
   \end{tabular}\\
    \end{tabular}
\end{minipage}
\hfill
  \begin{minipage}[t]{0.52\textwidth}  
      \vspace{0pt}
      \centering
        \small
      \renewcommand{\arraystretch}{1.4}
    \begin{tabular}{|>{\raggedright\arraybackslash}p{10em}>{\raggedright\arraybackslash}p{10em}|}
    \hline
    \includegraphics[scale=0.5]{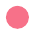}~LUNAR (\citeyear{woods1972lunar}) & \includegraphics[scale=0.5]{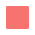}~CHAT-80 (\citeyear{warren-pereira-1982-efficient}) \\
\includegraphics[scale=0.5]{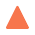}~PRECISE (\citeyear{popescu2003towards}) & \includegraphics[scale=0.5]{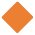}~SumTime (\citeyear{reiter2005choosing}) \\
\includegraphics[scale=0.5]{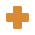}~\citet{pasupat-liang-2015-compositional} & \includegraphics[scale=0.5]{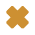}~Neural Enquirer (\citeyear{yin-etal-2016-neural}) \\
\includegraphics[scale=0.5]{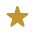}~Neural Programmer (\citeyear{neelakantan2017learning}) & \includegraphics[scale=0.5]{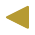}~\citet{wiseman-etal-2017-challenges} \\
\includegraphics[scale=0.5]{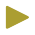}~\citet{liu2018table} & \includegraphics[scale=0.5]{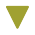}~Seq2SQL (\citeyear{zhong2018seqsql}) \\
\includegraphics[scale=0.5]{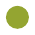}~SQLNet (\citeyear{xu2018sqlnet}) & \includegraphics[scale=0.5]{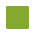}~TypeSQL (\citeyear{yu-etal-2018-typesql}) \\
\includegraphics[scale=0.5]{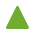}~SQLova (\citeyear{hwang2019comprehensive}) & \includegraphics[scale=0.5]{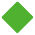}~RAT-SQL (\citeyear{wang-etal-2020-rat}) \\
\includegraphics[scale=0.5]{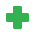}~RAT-SQL+ (\citeyear{wang-etal-2020-rat}) & \includegraphics[scale=0.5]{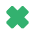}~RAT-SQL++ (\citeyear{wang-etal-2020-rat}) \\
 \includegraphics[scale=0.5]{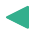}~TaBERT (\citeyear{yin-etal-2020-tabert}) &
\includegraphics[scale=0.5]{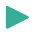}~Table-BERT (\citeyear{Chen2020TabFact}) \\
\includegraphics[scale=0.5]{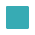}~\citet{chen-etal-2020-hybridqa} &
\includegraphics[scale=0.5]{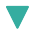}~TAPAS (\citeyear{herzig-etal-2020-tapas}) \\
\includegraphics[scale=0.5]{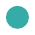}~TAPAS+ (\citeyear{herzig-etal-2020-tapas}) &
\includegraphics[scale=0.5]{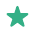}~GraPPa (\citeyear{yu2021grappa})\\
\includegraphics[scale=0.5]{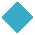}~GAP (\citeyear{shi2021learning}) &
\includegraphics[scale=0.5]{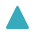}~\citet{chen2021open}\\
\includegraphics[scale=0.5]{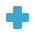}~UnifiedSKG (\citeyear{xie-etal-2022-unifiedskg}) &
\includegraphics[scale=0.5]{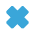}~UnifiedSKG+ (\citeyear{xie-etal-2022-unifiedskg}) \\ \includegraphics[scale=0.5]{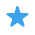}~PICARD (\citeyear{scholak-etal-2021-picard}) &
\includegraphics[scale=0.5]{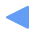}~TableFormer (\citeyear{yang-etal-2022-tableformer}) \\
 \includegraphics[scale=0.5]{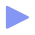}~TAPAX (\citeyear{liu2022tapex}) &
\includegraphics[scale=0.5]{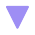}~TableLlama (\citeyear{zhang-etal-2024-tablellama}) \\ \includegraphics[scale=0.5]{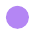} TableLLM-7B (\citeyear{zhang2024tablellm}) &
\includegraphics[scale=0.5]{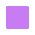} TableLLM-13B (\citeyear{zhang2024tablellm}) \\ \includegraphics[scale=0.5]{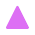} TableGPT-2 (\citeyear{su2024tablegpt2}) &
\includegraphics[scale=0.5]{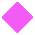} TableLlava-7B (\citeyear{zheng-etal-2024-multimodal}) \\
\includegraphics[scale=0.5]{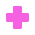}~TableLlava-13B (\citeyear{zheng-etal-2024-multimodal}) &
\includegraphics[scale=0.5]{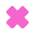}~TableBenchLLM-7B (\citeyear{wu2025tablebench})\\
\includegraphics[scale=0.5]{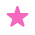}~TableBenchLLM-8B (\citeyear{wu2025tablebench})
&
\includegraphics[scale=0.5]{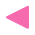}~TAMA (\citeyear{deng2025rethinking}) \\
\includegraphics[scale=0.5]{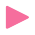}~Table-R1 (\citeyear{yang2025table}) & \\
\hline
\end{tabular}
\end{minipage}
\caption{Summarization of different eras for table modeling.
We note that the model sizes increase logarithmically with time.
When we enter the LLM era, the community has shifted its attention to instruction tune the foundational models \citep{zhang-etal-2024-tablellama}.
While there are persistent challenges, such as generalization for table models \citep{warren-pereira-1982-efficient, yu-etal-2018-spider, suhr-etal-2020-exploring, deng2025rethinking} across different eras, new challenges emerge. 
\Cref{app-subsec: table-model-history-fig-explain} provides additional details of this plot.
}
\label{fig: table-modeling-history}
\end{figure*}

%% file: figures/in-domain-avg.tex
\begin{figure*}[t]
    \centering
     \begin{subfigure}[t]{0.24\textwidth}
        \centering
        \includegraphics[width=\linewidth]{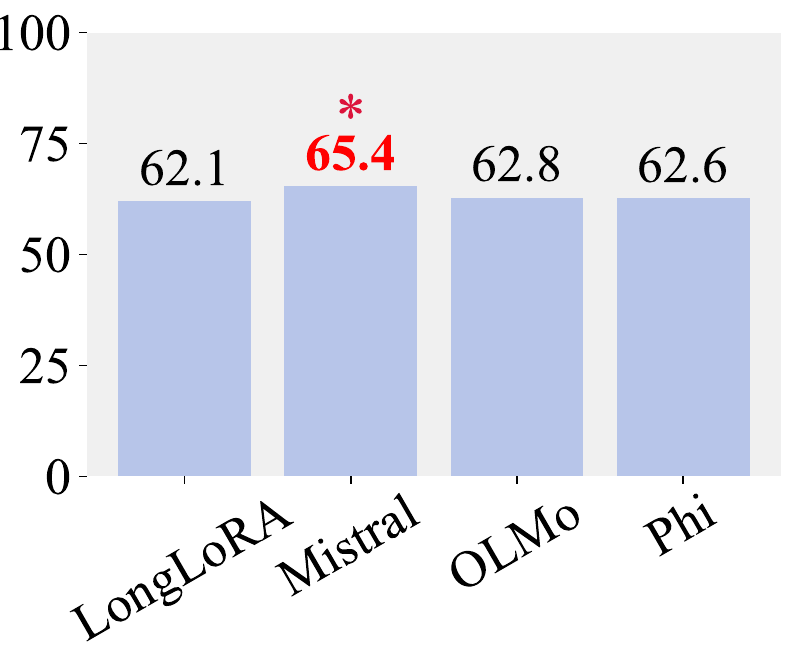}
        \caption{TableLlama.}
    \end{subfigure}%
     \begin{subfigure}[t]{0.24\textwidth}
        \centering
        \includegraphics[width=\linewidth]{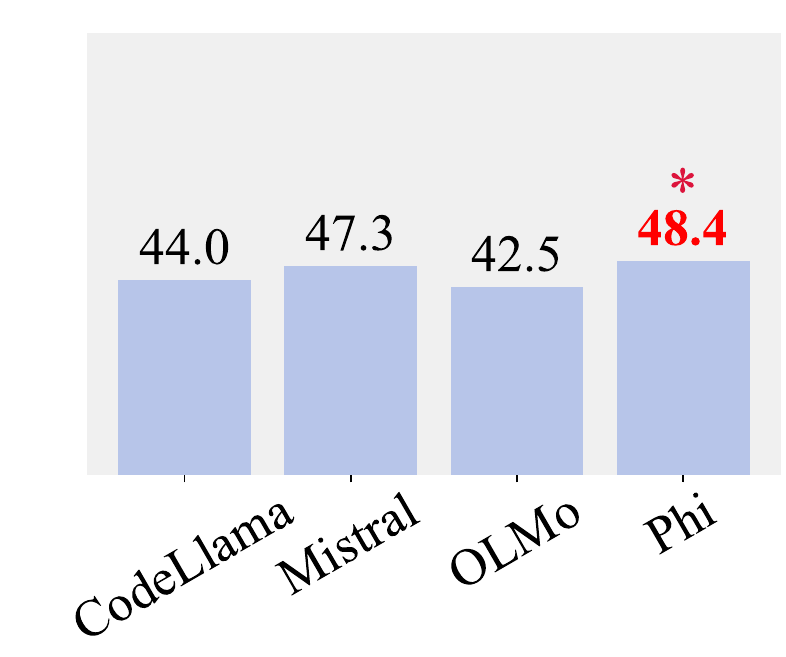}
        \caption{TableLLM.}
    \end{subfigure}%
     \begin{subfigure}[t]{0.24\textwidth}
        \centering
        \includegraphics[width=\linewidth]{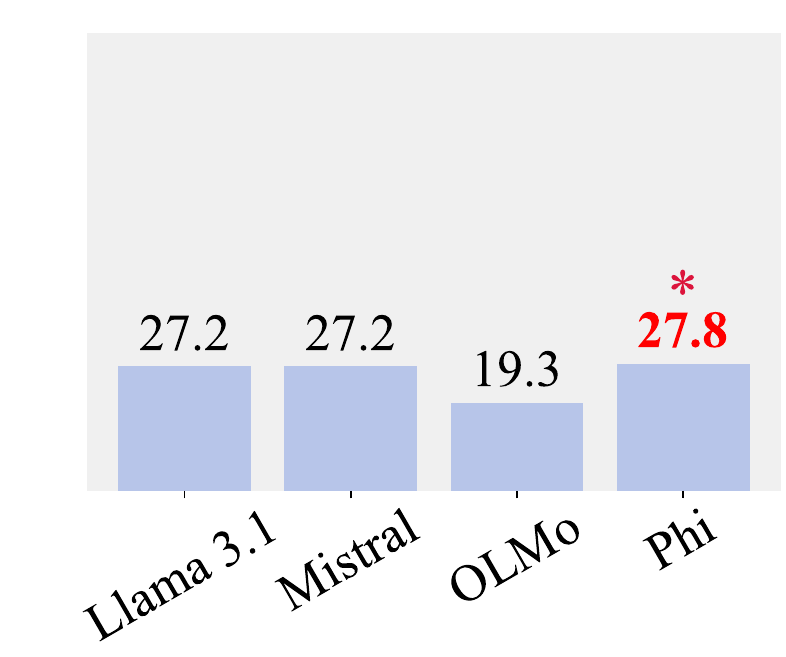}
        \caption{TableBench.}
    \end{subfigure}%
    \begin{subfigure}[t]{0.24\textwidth}
        \centering
        \includegraphics[width=\linewidth]{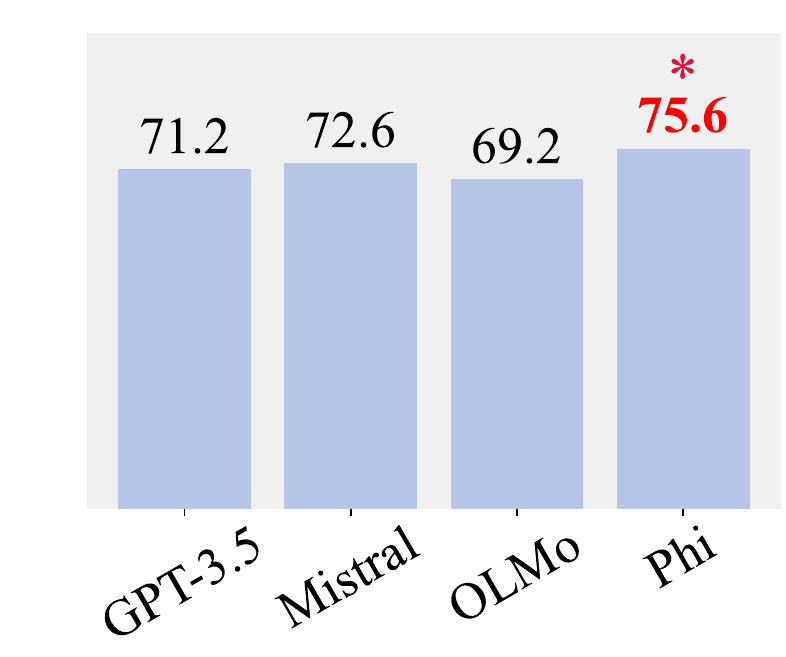}
        \caption{Table-GPT.}
    \end{subfigure}%

    \caption{Averaged in-domain performance (y-axis) between the models in existing works (the leftmost bar for each plot) versus our replications.
    Our replicated models achieve better in-domain results than the existing works.
    The detailed in-domain performance is reported in \Cref{app-sec: reimpl-compare}.
    }
    \label{fig:in-domain-perf}
\end{figure*}

%% file: tables/out-of-domain-complete.tex
\begin{table*}[!t]
\small
\centering
\renewcommand{\arraystretch}{1.3}
\setlength\tabcolsep{2pt}
\resizebox{\linewidth}{!}{
\begin{tabular}{rccccccccccccccccc}
\hline
\multicolumn{1}{c|}{\multirow{4}{*}{\textbf{\begin{tabular}[c]{@{}c@{}}Train\\Data\end{tabular}}}} & \multicolumn{8}{c|}{\textit{Real}} & \multicolumn{8}{c}{\textit{Synthesized}}\\
\cline{2-17}
& \multicolumn{5}{|c|}{\textbf{Table QA}} & \multicolumn{2}{c|}{\textbf{Fact Veri.}} & \multicolumn{1}{c|}{\textbf{Tab2Text}} & \multicolumn{7}{c|}{\textbf{Schema Reasoning}} &  \multicolumn{1}{c}{\textbf{Misc.}}\\
\cline{2-17}
& \multicolumn{1}{|c|}{FeT} & \multicolumn{1}{c|}{HiT}  & \multicolumn{1}{c|}{TabM} & \multicolumn{1}{c|}{TAT}  & \multicolumn{1}{c|}{Wiki}  & \multicolumn{1}{c|}{TabF} & \multicolumn{1}{c|}{Inf} & \multicolumn{1}{c|}{ToT} & \multicolumn{1}{c|}{Beer} & \multicolumn{1}{c|}{DeepM}  & \multicolumn{1}{c|}{DI} & \multicolumn{1}{c|}{ED}  & \multicolumn{1}{c|}{C}  & \multicolumn{1}{c|}{CF} & \multicolumn{1}{c|}{CTA} & \multicolumn{1}{c}{TabB\textsubscript{eval}} \\
 \cline{2-17}
& \multicolumn{1}{|c|}{BLEU} & \multicolumn{1}{c|}{Acc} & \multicolumn{1}{c|}{Acc} & \multicolumn{1}{c|}{Acc} & \multicolumn{1}{c|}{Acc} & \multicolumn{1}{c|}{Acc} & \multicolumn{1}{c|}{Acc} & \multicolumn{1}{c}{BLEU} & \multicolumn{1}{|c|}{F1} & \multicolumn{1}{c|}{Recall} & \multicolumn{1}{c|}{Acc} & \multicolumn{1}{c|}{F1} & \multicolumn{1}{c|}{F1} & \multicolumn{1}{c|}{Acc} & \multicolumn{1}{c|}{F1} & \multicolumn{1}{c}{ROUGE-L}\\
\hline
\multicolumn{17}{l}{\grayc Mistral v0.3 7B Instruct} \\
$\crown$ N/A & \textbf{20.0} & 35.5 & 66.9 & 18.0 & 27.9 & \textbf{62.3} & \textbf{42.8} & 11.5 & \red\textbf{97.2} & 42.9 & 27.9 & 24.1 & \textbf{30.2} & 19.1 & 63.8 & 18.9 \\
TableLlama & \g 38.7 & \g 70.6 & 71.2 & 5.6 & 23.8 & \g86.8 & 27.7 & \textbf{28.5} & 25.8 & 70.0 & 13.4  & 25.1 & 17.4 & \hspace{0.5em}0.5  & 34.9 & \textbf{19.6}\\
$\crown$ TableLLM & 10.2 & \textbf{44.1} & \textbf{75.0} & 25.0 & 32.3 & 11.9 & 15.4 & 6.7 & 45.0 & \red\textbf{78.6} & \textbf{33.1} & \textbf{43.1} & 25.6 & 15.0 &  66.9 & 3.7 \\
TableBench & 7.9 & \textbf{44.1} & 70.6 & \red\textbf{25.7} & \textbf{37.4} & 36.5 & 27.5 & 3.5 & 88.5 & 50.0 & 32.0 & 20.3 & 27.4 & 13.3 & \textbf{72.2} & \g27.2\\
TableGPT & 19.5 & 35.8 & 62.2 & 14.1 & 25.5 & 61.4 & 35.8 & 4.5 & \g100.0 & \g98.0 & \g46.4 & \g46.0 & 23.8 & \textbf{25.3} & 68.3 & 13.1 \\ 
\hline
\multicolumn{17}{l}{\grayc OLMo 7B Instruct} \\
N/A & 6.0 & 27.3 & 54.4 & 14.3 & 19.4 & 38.2 & 21.4 & 5.1 & \textbf{50.5} & 35.7 & 28.9 & 14.1 & 15.0 & 16.2 & 54.5 & 7.6\\
TableLlama & \g36.8 & \g67.9 & \textbf{72.9} & 9.9 & 6.7 & \g83.8 & 15.0 & \textbf{20.8} & 0.0 & 7.1 & 21.2 & 14.6 & 14.8 & 10.7 & 23.5 & \textbf{17.1}\\
$\crown$ TableLLM & \textbf{9.7} & \textbf{35.5} & 65.5 & \textbf{17.7} & 26.7 & 40.6 & 16.9 & 8.9 & 16.5 & \textbf{42.9} & 33.0 & \textbf{37.6} & 13.0 & 18.7 & 43.6 & 6.3\\
TableBench & 3.8 & 28.3 & 62.6 & 15.6 & \textbf{34.0} & 30.9 & 6.5 & 7.5 & 43.4 & 16.6 & \red\textbf{36.6} & 28.6 & 18.1 & 21.2 & 46.5 & \g19.3 \\
$\crown$ TableGPT & 9.3 & 27.2 & 65.6 & 14.6 & 16.4 & \textbf{44.9} & \textbf{33.0} & 11.4 & \g96.2 & \g100.0 & \g45.4 & 35.3  & \textbf{19.9} & \textbf{29.3} & \textbf{62.5}  & 13.7 \\
\hline 
\multicolumn{17}{l}{\grayc Phi 3 Small Instruct (7B)} \\
N/A & 5.0 & 39.6 & 76.1 & 13.0 & 29.7 & 65.3 & 62.3 & 1.4  & 95.0 & 42.9 & 31.9 & 49.7 & 30.6 & 43.4 & 71.5 & 8.3 \\
TableLlama & \g38.1 & \g63.6 & 74.8 & 18.3 & \red\textbf{46.3} & \g86.2 & 54.3 & \red\textbf{29.6}  & \textbf{95.6} & 35.7 & 4.3 & 19.4 & 27.9 & 36.5 & 43.9 & 22.4\\
$\crown$ TableLLM & 18.2 & \red\textbf{45.3} & 81.2 & \textbf{24.1} & 37.7 & 69.6 & 44.6 & 8.1  & 80.2 & \textbf{50.0} & 34.0 & 41.3 & 27.9 & \red\textbf{49.5} & 70.1 & \red\textbf{27.2} \\
$\crown$ TableBench & 10.0 & 3.5 & \red\textbf{83.0} & 20.5 & 34.6 & 68.0 & 65.3 & 0.9 & 95.0 & 28.6 & \textbf{35.9} & \red\textbf{53.8} & \red\textbf{31.1} & 46.2 & \red\textbf{76.7} & \g27.8\\
TableGPT & \red\textbf{24.8} & 45.1 & 76.8 & 15.6 & 30.0 & \red\textbf{71.0} & \red\textbf{67.0} & 14.0 & \g98.9 & \g98.8 & \g49.4 & \g55.4 & 24.8 & 45.2 & 68.3 & 26.1 \\
\hline
\end{tabular}
}
\caption{Evaluation for table tasks.
Gray indicates that the model is trained on the corresponding training set.
Bolded numbers represent the best performance among variants of the same base model, while red is the best overall performance across all models.
Mistral v0.3 7B Instruct, OLMo 7B Instruct, and Phi 3 Small Instruct (7B) indicate the base model on which we apply the training data, respectively.
``$\crown$'' marks the model that has the most number of top performance across all the datasets with respect to the same base model.
We note that Phi-based models yield the highest performance scores across most of the out-of-domain table datasets, while TableLLM training data consistently yield the most top performance across different base LLMs.
}
\label{tab: out-of-domain-complete}
\end{table*}

%% file: figures/general-benchmark-performance-difference.tex
\begin{figure*}[t]
    \centering
    \begin{subfigure}[t]{0.9\linewidth}
        \centering
        \includegraphics[width=\linewidth]{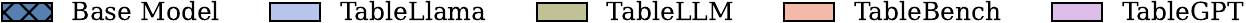}
    \end{subfigure}%
    \vspace*{0.4em}
     \begin{subfigure}[t]{0.32\textwidth}
        \centering
        \includegraphics[width=\linewidth]{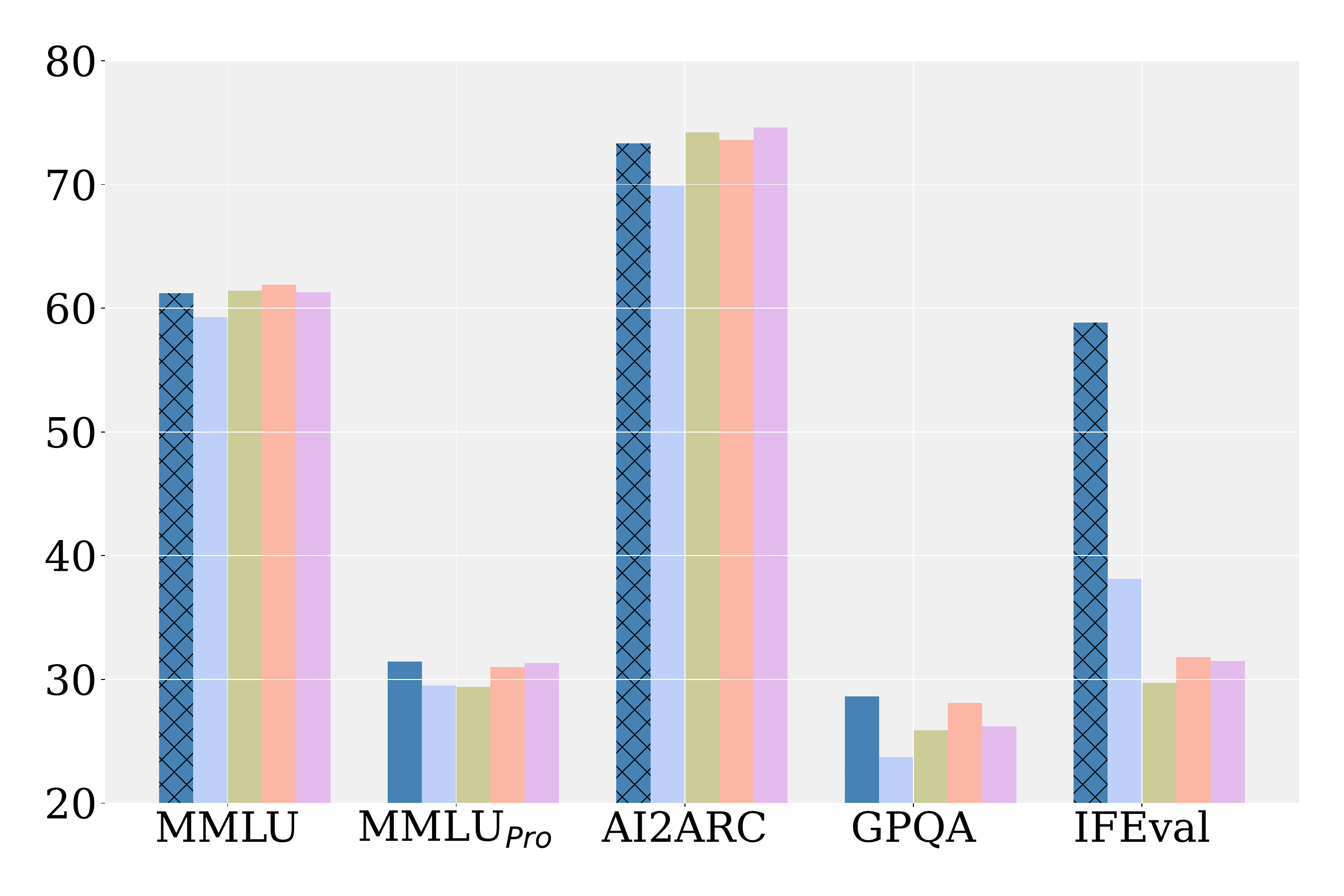}
        \caption{Mistral v0.3 7B Instruct.}
    \end{subfigure}%
     \begin{subfigure}[t]{0.32\textwidth}
        \centering
        \includegraphics[width=\linewidth]{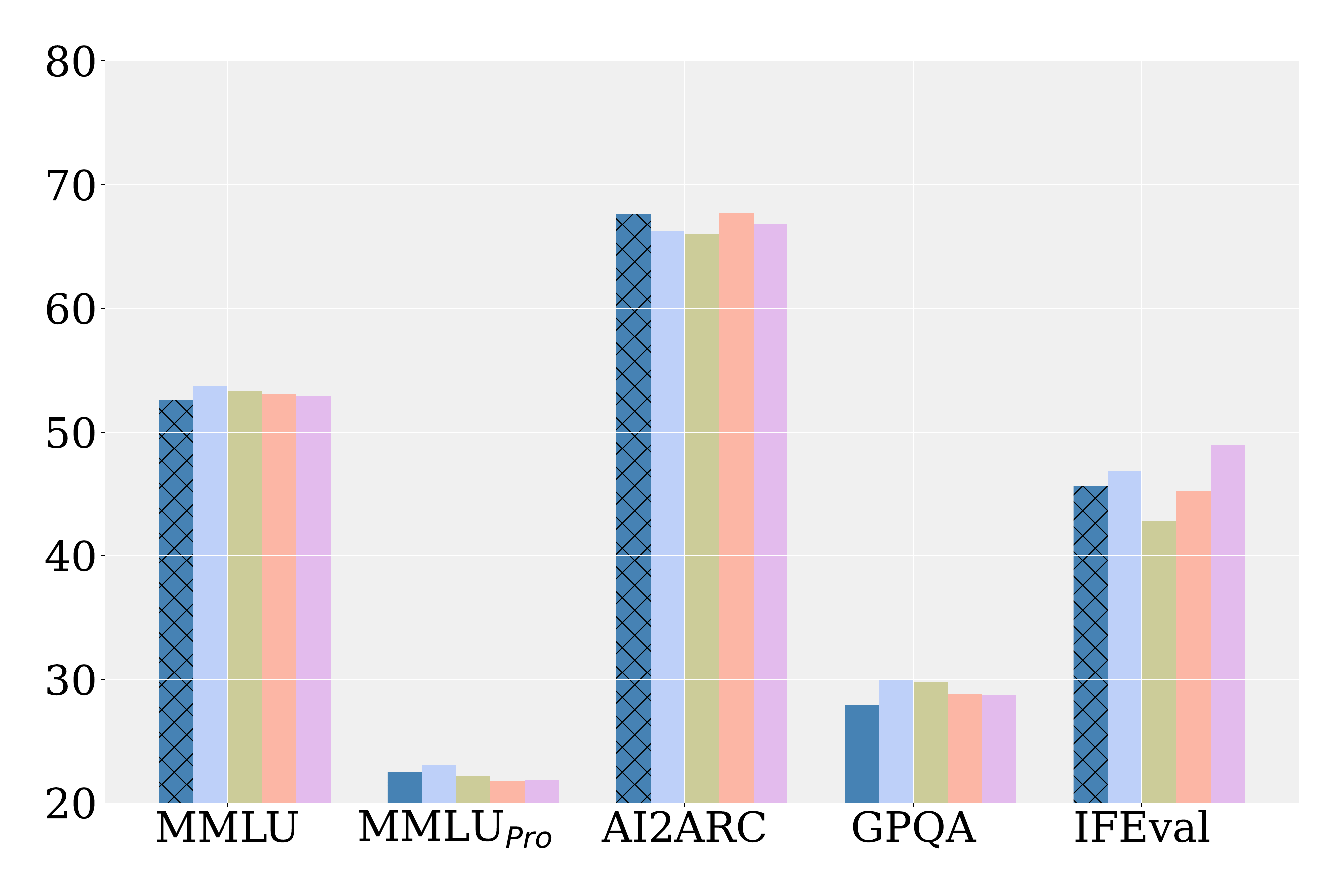}
        \caption{OLMo 7B Instruct.}
    \end{subfigure}%
     \begin{subfigure}[t]{0.32\textwidth}
        \centering
        \includegraphics[width=\linewidth]{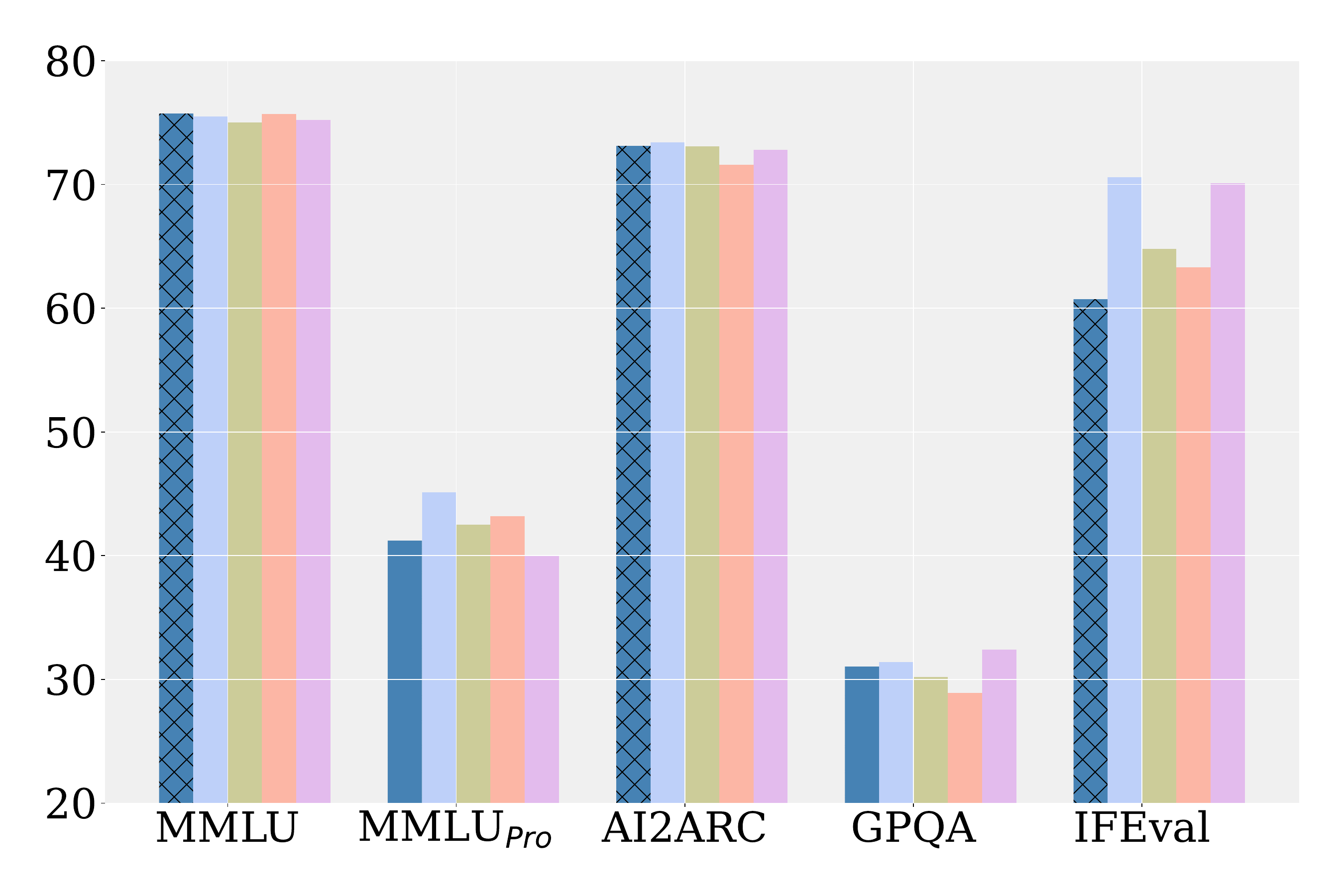}
        \caption{Phi 3 Small Instruct (7B).}
    \end{subfigure}%

    \caption{Performance of fine-tuned models trained on different data (e.g. TableLlama) on general benchmarks. 
    The green and red hatched bars represent performance gains or losses relative to the base model, respectively.
    As indicated by the similar performance bar heights, table instruction tuning does not necessarily compromise the base model's general capabilities.
    \Cref{tab:general} provides the performance in number.
    }
    \label{fig:general-benchmark-performance-difference}
    \vspace{-0.5em}
\end{figure*}